\documentclass[10pt,twocolumn,letterpaper]{article}

\usepackage[pagenumbers]{cvpr} 

\usepackage[accsupp]{axessibility}  
\usepackage{comment}
\usepackage{multirow}
\usepackage{balance}
\usepackage{flushend}
\usepackage{colortbl}
\definecolor{Gray}{gray}{0.93}
\definecolor{lightblue}{rgb}{0.90, 0.95, 0.99}

\usepackage{amsthm}

\theoremstyle{remark}

\usepackage{upgreek}

\newcommand{\name}{VecAttention}

\usepackage{bm}
\usepackage{algorithm}
\usepackage{algorithmic}
\definecolor{comment_color_2}{RGB}{64,128,128}
\definecolor{comment_color}{HTML}{AE4132}
\newcommand{\LineComment}[1]{\vspace*{0.5em}\small\textcolor{comment_color_2}{\textit{\# #1}}}
\newcommand{\ParallelComment}[1]{\vspace*{0.5em}\small\textcolor{comment_color}{\textit{\# #1}}}

\DeclareMathOperator{\Idx}{Idx}

\definecolor{cvprblue}{rgb}{0.21,0.49,0.74}
\usepackage[pagebackref,breaklinks,colorlinks,allcolors=cvprblue]{hyperref}

\def\paperID{} %
\def\confName{CVPR}
\def\confYear{2026}

\title{\name{}: Vector-wise Sparse Attention for Accelerating Long Context Inference}

\author{
    Anmin Liu\textsuperscript{1,2}\thanks{Work done during internship at Alibaba Group.},\hspace{0.1cm}
    Ruixuan Yang\textsuperscript{3},\hspace{0.1cm}
    Huiqiang Jiang\textsuperscript{4},\hspace{0.1cm}
    Bin Lin\textsuperscript{4},\hspace{0.1cm}
    Minmin Sun\textsuperscript{4},\\
    Yong Li\textsuperscript{4},\hspace{0.1cm}
    Chen Zhang\textsuperscript{5,}\thanks{Corresponding authors.},\hspace{0.1cm}
    Tao Xie\textsuperscript{2,1,}\footnotemark[2]\\
    \textsuperscript{1}SCS, Peking University, Beijing, China \quad 
    \textsuperscript{2}Key Lab of HCST (PKU), MOE, Beijing, China \\
    \textsuperscript{3}Fudan University, Shanghai, China \quad
    \textsuperscript{4}Alibaba Group, China \\
    \textsuperscript{5}Shanghai Jiao Tong University, Shanghai, China
}

\begin{document}
\maketitle
\begin{abstract}
Long-context video understanding and generation pose a significant computational challenge for Transformer-based video models due to the quadratic complexity of self-attention. While existing sparse attention methods employ coarse-grained patterns to improve efficiency, they typically incur redundant computation and suboptimal performance. To address this issue, in this paper, we propose \textbf{\name{}}, a novel framework of vector-wise sparse attention that achieves superior accuracy-efficiency trade-offs for video models. We observe that video attention maps exhibit a strong vertical-vector sparse pattern, and further demonstrate that this vertical-vector pattern offers consistently better accuracy-sparsity trade-offs compared with existing coarse-grained sparse patterns. Based on this observation, \name{} dynamically selects and processes only informative vertical vectors through a lightweight important-vector selection that minimizes memory access overhead and an optimized kernel of vector sparse attention. Comprehensive evaluations on video understanding (VideoMME, LongVideoBench, and VCRBench) and generation (VBench) tasks show that VecAttention delivers a 2.65$\times$ speedup over full attention and a 1.83$\times$ speedup over state-of-the-art sparse attention methods, with comparable accuracy to full attention. Our code is available at \url{https://github.com/anminliu/VecAttention}.
\end{abstract}    
\section{Introduction}
\label{sec:intro}

Large video models have recently advanced rapidly across two major tracks: video understanding (Vision-Language Models in short as VLMs)~\cite{zhang2024vision,chen2024videollm,chen2024longvila} and video generation (Diffusion Transformers in short as DiTs)~\cite{peebles2023scalable,kong2024hunyuanvideo,wan2025wan}. However, the quadratic complexity of attention with respect to the context length makes inference latency grow sharply with longer videos. In practical deployments, this factor leads to latency at the minute to hour level for realistic inputs~\cite{li2025mminference,xi2025sparse}, inflating serving cost and limiting scalability in production.
A growing body of evidence~\cite{li2025mminference,jiang2024minference,xu2025xattention,lai2025flexprefill} suggests that attention maps are inherently sparse: only a small subset of token-to-token interactions substantially contributes to model outputs (we refer to the ideal interaction selection as the oracle sparse pattern). Leveraging sparsity can reduce FLOPs without sacrificing quality. 
\begin{figure}[t]
    \centering
    \begin{subfigure}[b]{\linewidth}
        \centering
        \includegraphics[width=\textwidth]{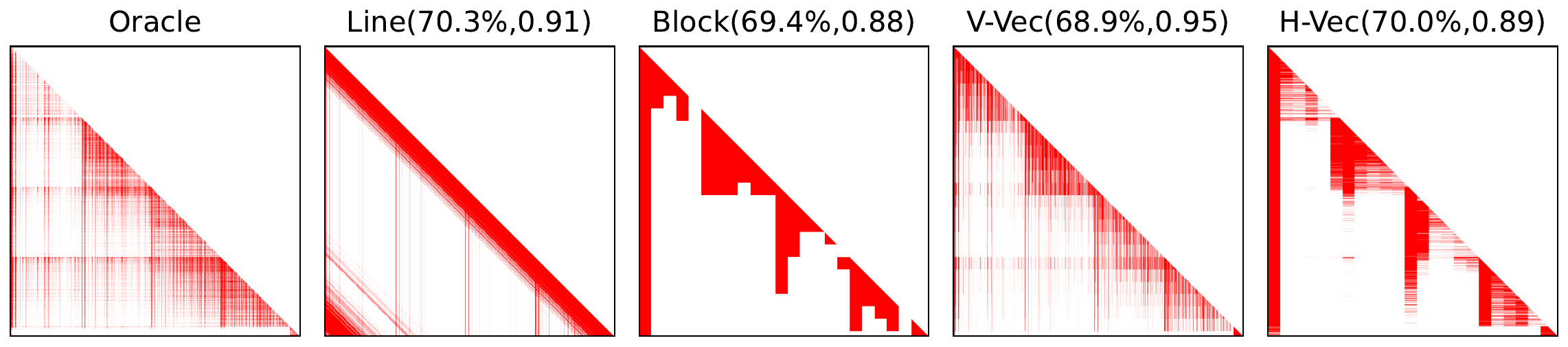}
        \caption{Video understanding (causal VLM)}
        \label{fig:sparsePattern:a}
    \end{subfigure}
    \begin{subfigure}[b]{\linewidth}
        \centering
        \includegraphics[width=\textwidth]{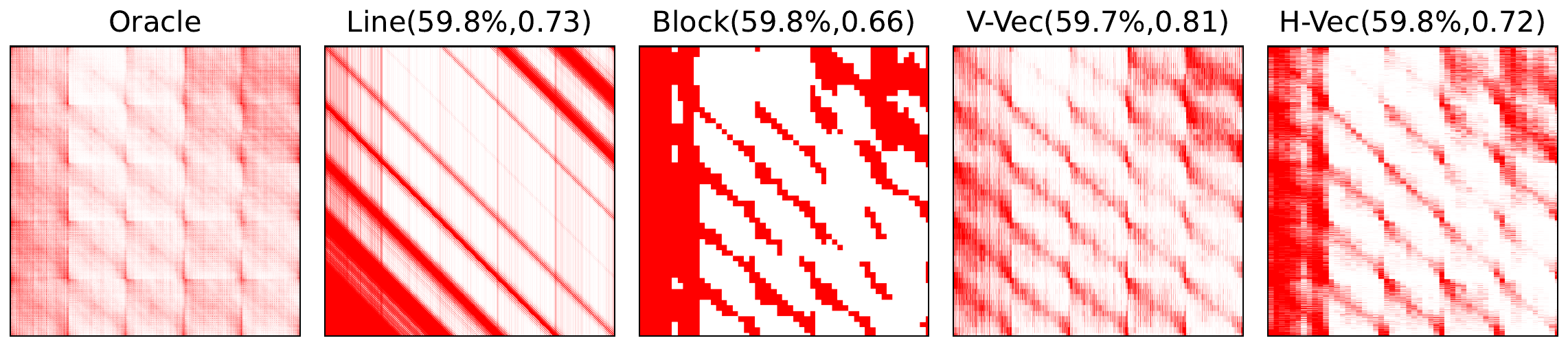}
        \caption{Video generation (non-causal DiT)}
        \label{fig:sparsePattern:b}
    \end{subfigure}
    \begin{subfigure}[b]{\linewidth}
        \centering
        \includegraphics[width=\textwidth]{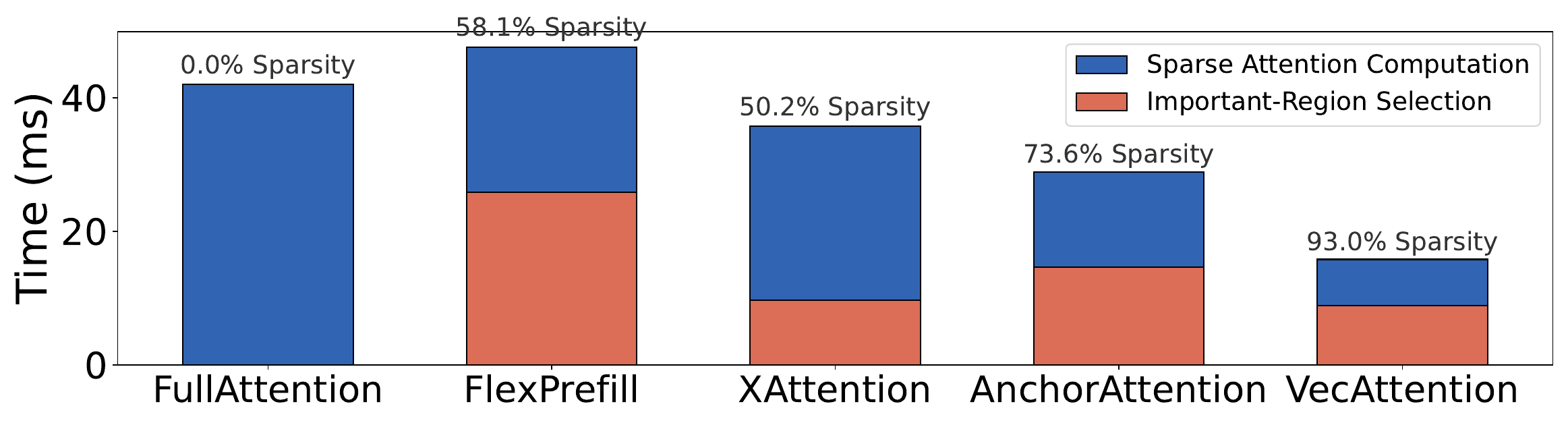}
        \caption{Attention breakdown}
        \label{fig:sparsePattern:c}
    \end{subfigure}
    \caption{
    (a–b) Under the same sparsity budget, \textbf{vertical-vector (V-Vec)} achieves higher recall and better approximates the oracle pattern than horizontal-vector (H-Vec) and other coarse line/block patterns, for both VLMs (video understanding) and DiTs (video generation). Markers such as (68.9\%, 0.95) denote (sparsity, recall), with sparsity in percentage.
    (c) On VideoMME at matched full-accuracy settings, \name{} attains higher effective sparsity and faster attention computation, with low important-region selection overhead, compared to existing coarse-grained methods.
    }
    
    \label{fig:sparsePattern}
\end{figure}

Existing work typically imposes coarse-grained patterns, most notably (1) line patterns, e.g., grid-like or vertical-slash strided selections~\cite{li2025mminference,jiang2024minference,lai2025flexprefill}, and (2) block patterns, which activate a few rectangular blocks with high estimated importance~\cite{xu2025xattention,guo2024blocksparse,zhang2025spargeattn}. Such work indeed cuts global computation, but often keeps redundant computation inside each selected region, leaving a sizable gap to the oracle.

We empirically observe that neither line nor block masks from the existing work closely approximate the oracle in recall (how informative the preserved entries are) or sparsity (how many entries are dropped), as shown in \Cref{fig:sparsePattern:a} and \Cref{fig:sparsePattern:b}. More importantly, their granularity is mismatched to how information concentrates in video attention: the most salient entries are typically organized into fine, fragmented line segments, rather than forming continuous straight lines traversing the entire attention map or bulky rectangular regions. This mismatch explains why coarse masks still compute many uninformative interactions. 
We further validate this observation in \Cref{sec:observation}, where experiments across diverse video scenarios show that vertical-vector sparsity provides a far closer approximation to the oracle than coarse line or block patterns. 

To address the issue of the existing work, in this paper, we propose \name{}, a framework of fine-grained vector-wise sparse attention tailored for long-context video inference in both VLMs and DiTs. \name{} operationalizes the preceding observation in two steps. 
(1) Granularity selection. We define the region unit as a column vector in the attention map (a “vertical vector”), which aligns with temporal query similarity and naturally trades off accuracy vs. sparsity at a finer granularity than line/block masks.
(2) Efficient important-vector selection. Fine granularity naively implies a large estimated attention map and prohibitive HBM load/store if one were to materialize it. \name{} avoids this issue by integrating on-the-fly selection into the tiled QK GEMM operation: each on-chip tile computes cheap statistics that identify informative columns without writing intermediate QK output tiles back to HBM. Thus, we can obtain the indices of columns with low extra overhead (see \Cref{fig:sparsePattern:c}), and then restrict subsequent attention operations to those columns.

Across diverse video lengths and resolutions, \name{} consistently reduces inference latency without undermining accuracy, demonstrating that fine-grained vector-wise sparsity is a practical and accurate alternative to coarse line/block patterns for large video models. Specifically, experiments across multiple benchmarks of video understanding and generation show that \name{} can achieve a 2.65$\times$ speedup over full attention and a 1.83$\times$ speedup over state-of-the-art sparse attention methods, while maintaining accuracy comparable to full attention.

\section{Background and Observation}
\label{sec:observation}
\subsection{Attention Formulation}
We first introduce the computation of the attention mechanism in video models as follows:
\begin{equation}
\label{eq:attention_map}
\boldsymbol{S} = \frac{\boldsymbol{Q} \boldsymbol{K}^\top}{\sqrt{D}} ,\,
\boldsymbol{A} = \operatorname{softmax}\!\left(\boldsymbol{S}\right),\,
\boldsymbol{O} = \boldsymbol{A}\boldsymbol{V},
\end{equation}
where 
$\boldsymbol{Q} = [\boldsymbol{q}_1,\ldots,\boldsymbol{q}_N]^\top \in \mathbb{R}^{N\times D}$, 
$\boldsymbol{K} = [\boldsymbol{k}_1,\ldots,\boldsymbol{k}_N]^\top \in \mathbb{R}^{N\times D}$, 
and $\boldsymbol{V} = [\boldsymbol{v}_1,\ldots,\boldsymbol{v}_N]^\top \in \mathbb{R}^{N\times D}$ 
denote the query, key, and value matrices, respectively. 
Here $N$ denotes the input context length, $\boldsymbol{S}$ denotes the attention scores,  
$\boldsymbol{A} \in \mathbb{R}^{N\times N}$ represents the attention map whose element $\boldsymbol{A}[i,j]$ measures
the attention weight between $\boldsymbol{q}_i$ and $\boldsymbol{k}_j$, and 
$\boldsymbol{O} \in \mathbb{R}^{N\times D}$ is the attention output.

\begin{figure}[t]
    \centering
    \begin{subfigure}[b]{\linewidth}
        \centering
        \includegraphics[width=\textwidth]{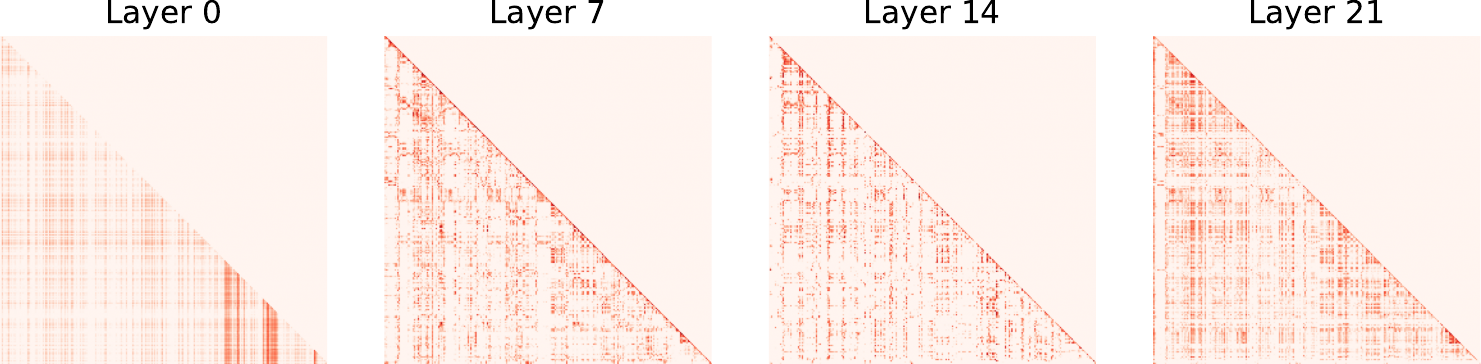}
        \caption{Video Understanding}
    \end{subfigure}
    \begin{subfigure}[b]{\linewidth}
        \centering
        \includegraphics[width=\textwidth]{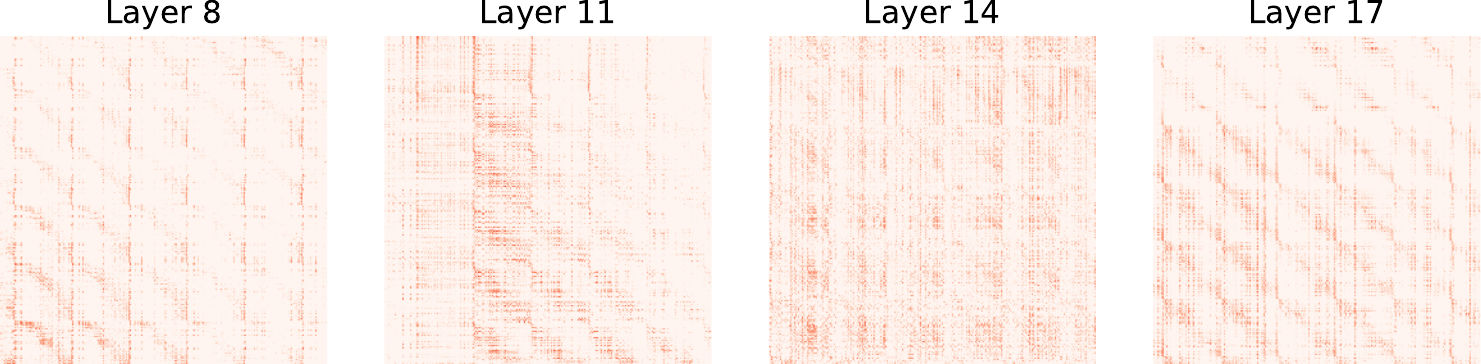}
        \caption{Video Generation}
    \end{subfigure}
    \caption{
    Attention maps visualized on different layers across tasks of video understanding and video generation.
    }
    \label{fig:visualizationMap}
\end{figure}

\begin{figure*}[t]
    \centering

    \begin{subfigure}[b]{0.235\textwidth}
        \centering
        \includegraphics[width=\textwidth]{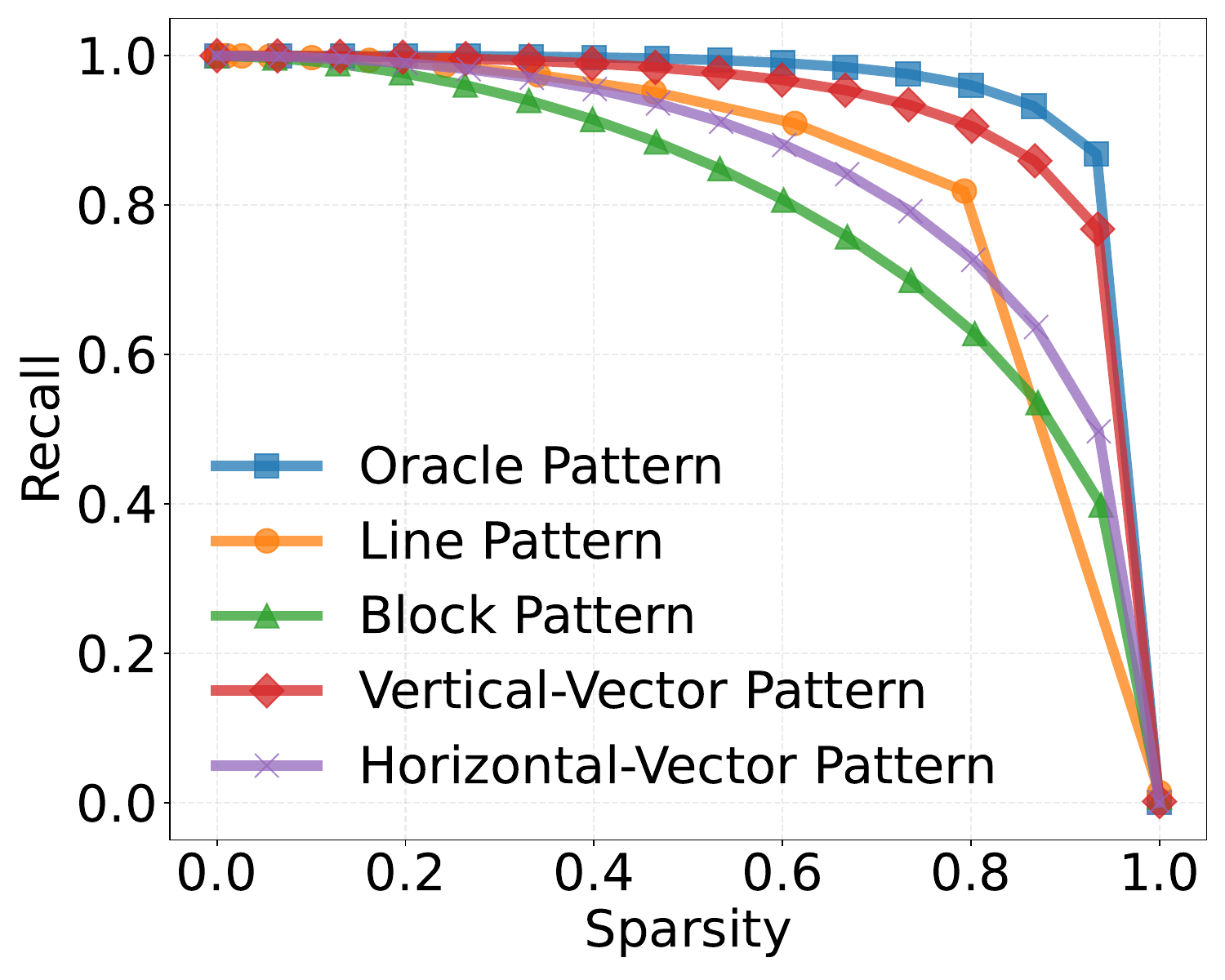}
        \caption{VideoMME (InternVL)}
        \label{fig:speedup_acc_A}
    \end{subfigure}
    \hfill
    \begin{subfigure}[b]{0.235\textwidth}
        \centering
        \includegraphics[width=\textwidth]{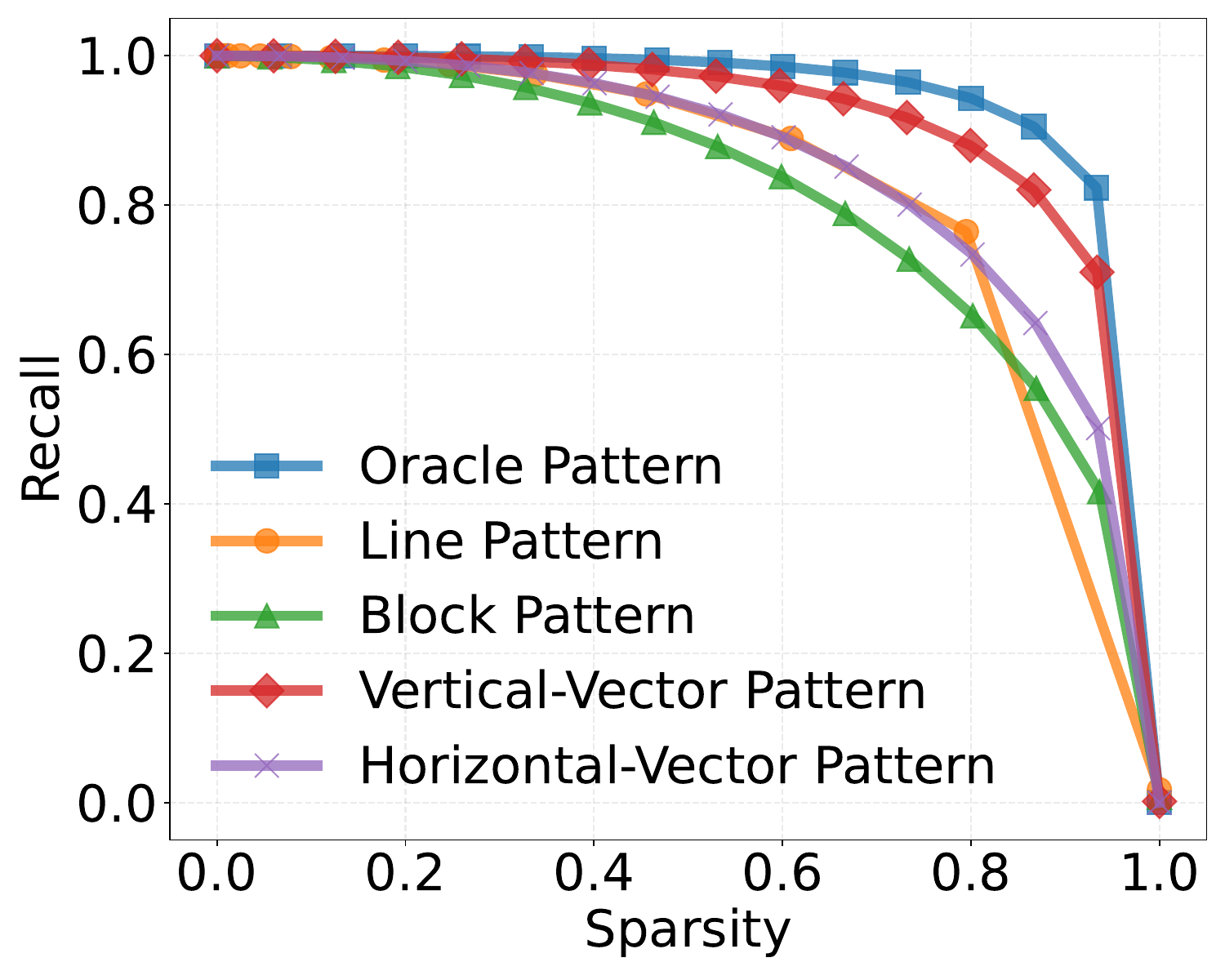}
        \caption{LongVideoBench (InternVL)}
        \label{fig:speedup_acc_B}
    \end{subfigure}
    \hfill
    \begin{subfigure}[b]{0.235\textwidth}
        \centering
        \includegraphics[width=\textwidth]{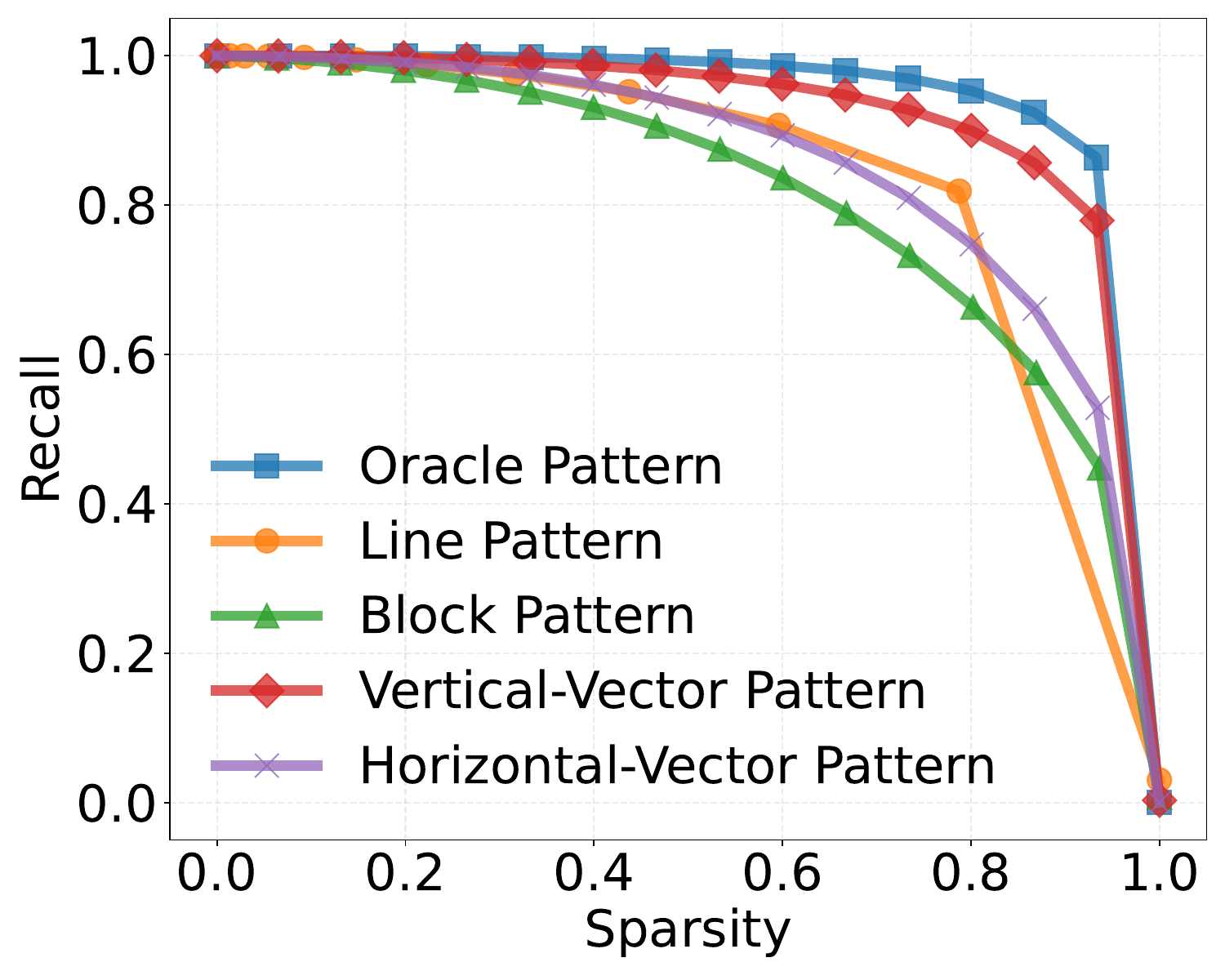}
        \caption{VCRBench (InternVL)}
        \label{fig:speedup_acc_C}
    \end{subfigure}
    \hfill
    \begin{subfigure}[b]{0.235\textwidth}
        \centering
        \includegraphics[width=\textwidth]{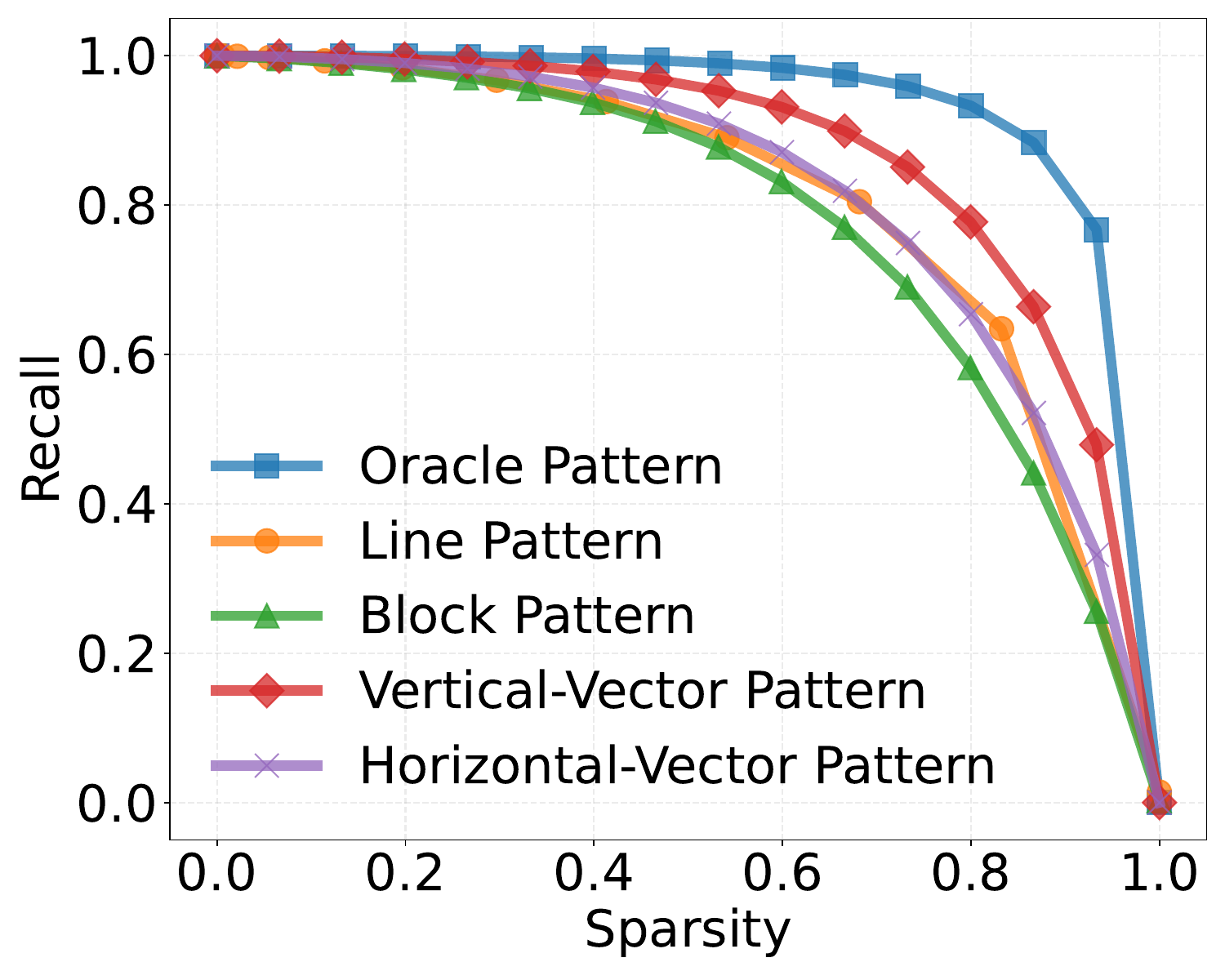}
        \caption{VBench (Wan)}
        \label{fig:speedup_acc_D}
    \end{subfigure}

    \begin{subfigure}[b]{0.235\textwidth}
        \centering
        \includegraphics[width=\textwidth]{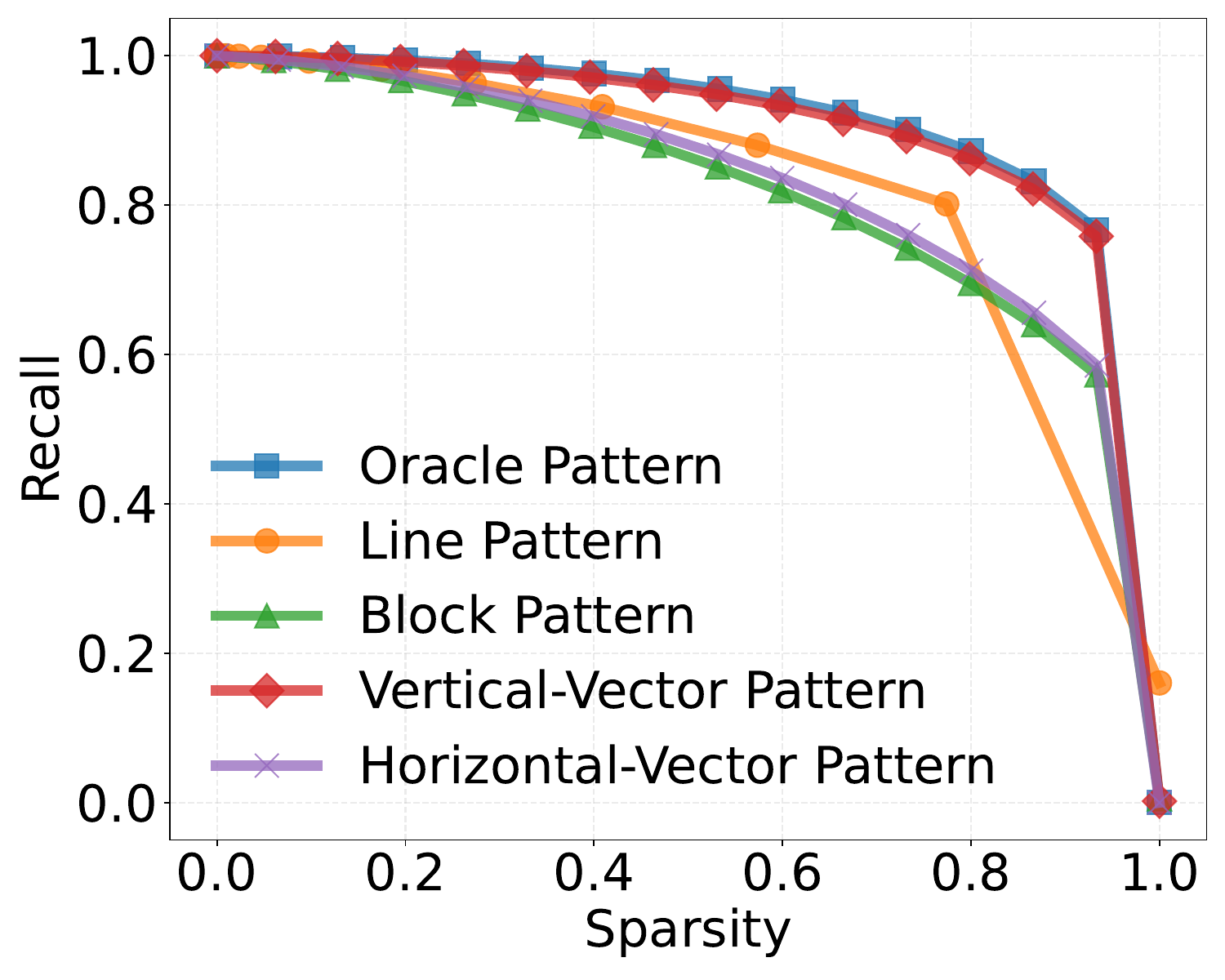}
        \caption{VideoMME (QwenVL)}
        \label{fig:sparsity_acc_A}
    \end{subfigure}
    \hfill
    \begin{subfigure}[b]{0.235\textwidth}
        \centering
        \includegraphics[width=\textwidth]{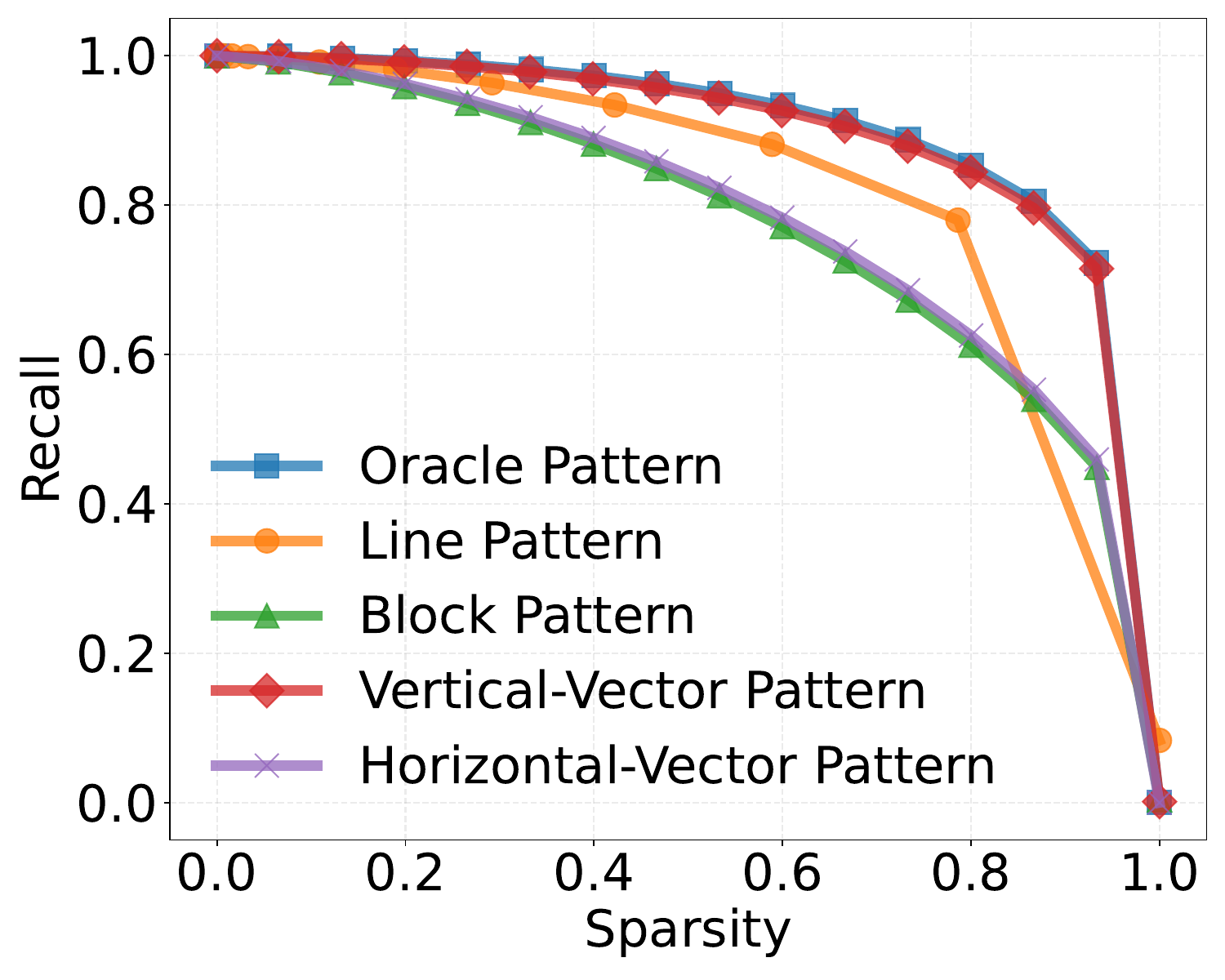}
        \caption{LongVideoBench (QwenVL)}
        \label{fig:sparsity_acc_B}
    \end{subfigure}
    \hfill
    \begin{subfigure}[b]{0.235\textwidth}
        \centering
        \includegraphics[width=\textwidth]{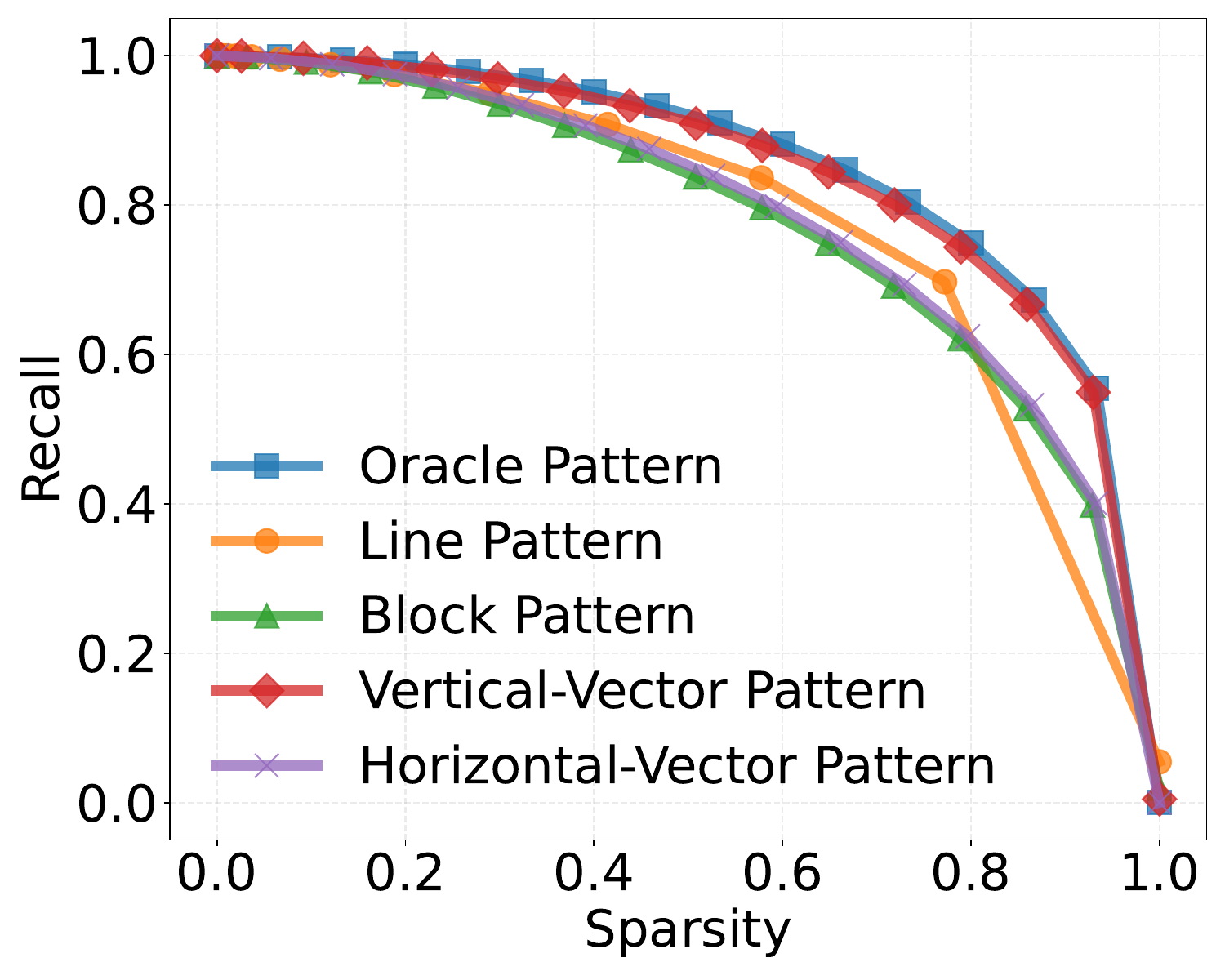}
        \caption{VCRBench (QwenVL)}
        \label{fig:sparsity_acc_C}
    \end{subfigure}
    \hfill
    \begin{subfigure}[b]{0.235\textwidth}
        \centering
        \includegraphics[width=\textwidth]{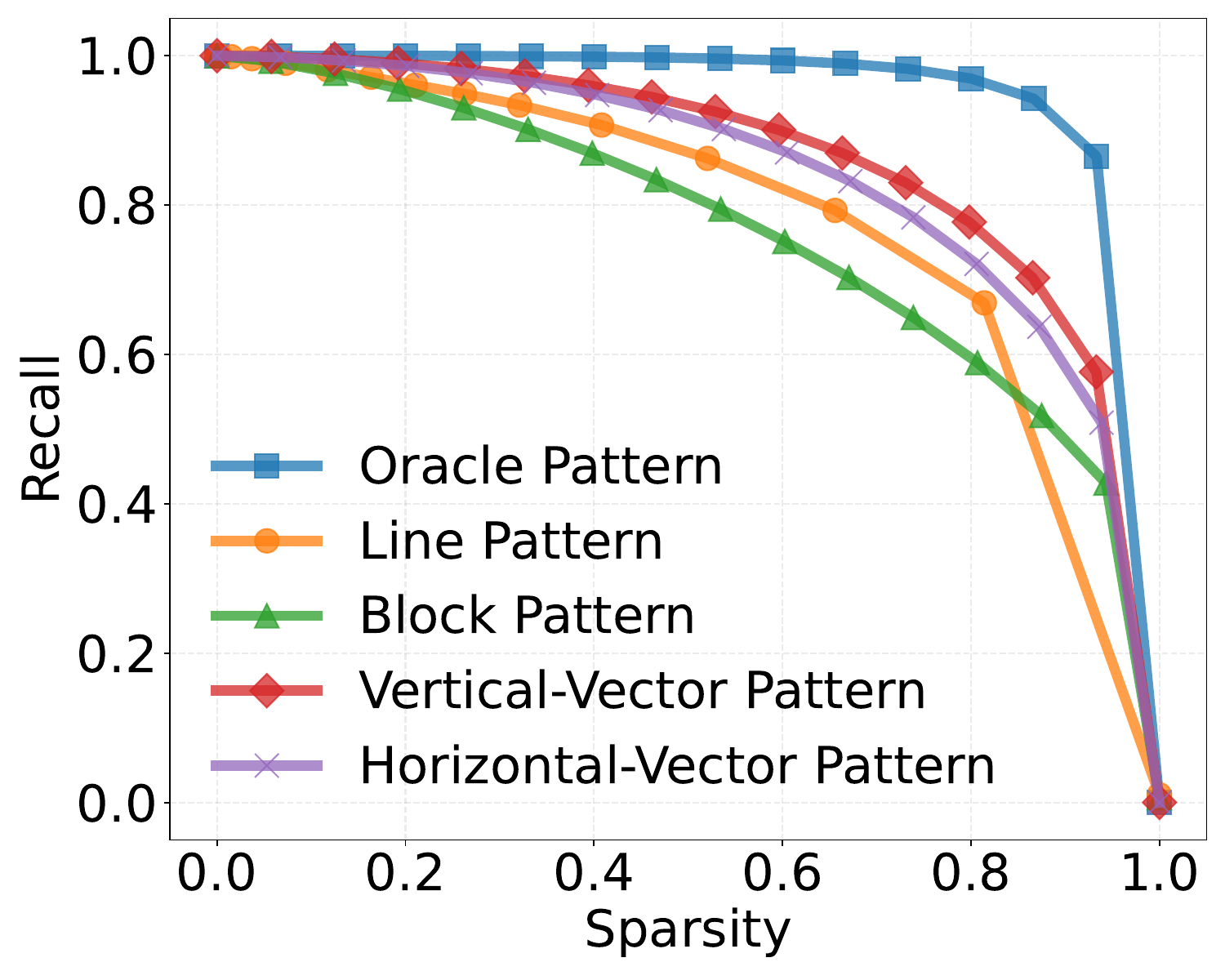}
        \caption{VBench (HunyuanVideo)}
        \label{fig:sparsity_acc_D}
    \end{subfigure}

    \caption{
    Sparsity--recall tradeoff of different sparse patterns across long-context video understanding (left three columns) and video generation (right column) tasks. The top row shows results on the causal InternVL-3.5-8B VLM and non-causal HunyuanVideo DiT, while the bottom row presents the causal Qwen2.5-VL-7B VLM and non-causal Wan2.1-T2V-14B DiT. The vertical-vector pattern consistently yields higher recall under the same sparsity than horizontal-vector pattern and coarse line/block pattern, closely tracking the oracle pattern across different datasets and models.
    }
    \label{fig:pattern_comparison_grid}
\end{figure*}

\subsection{Attention Exhibits Fine-Grained Vertical-Vector Pattern}
\textbf{Visualization of Attention Map.}
As shown in \Cref{fig:visualizationMap}, we visualize the attention maps of different video models across tasks of both video understanding and video generation. We first observe that the attention maps exhibit extremely high sparsity—most values are close to zero—indicating that the majority of computations in attention can be safely skipped, focusing on only a small set of informative regions. Furthermore, the sparse patterns (i.e., the shapes and spatial distributions of important regions) appear as fine-grained \textit{vertical vectors} scattered throughout the attention map. In contrast, existing coarse-grained sparse patterns~\cite{xu2025xattention,jiang2024minference} include many uninformative vertical vectors, leading to redundant computation. We provide a more comprehensive visualization exhibited in the attention maps in \Cref{sec:visualization}.

\textbf{Statistical Comparison of Different Sparse Patterns.}
\Cref{fig:pattern_comparison_grid} presents a statistical comparison of different sparse patterns in approximating the attention map.  
We evaluate all patterns under the following setting: the block size and vector size are both set to 128, the oracle pattern is regarded as a block pattern with block size~1, and the line pattern includes both vertical and slash lines.
For each pattern, we partition the full attention map into regions based on its granularity, compute region importance as the mean attention weights within each region, rank all regions in descending importance, and progressively select them.  
After each selection step, we update the mask and record the resulting sparsity and recall, yielding a sparsity-recall trade-off curve for each pattern.
We observe that, across different vision tasks and vision models, the fine-grained \textit{vertical-vector} sparse pattern consistently outperforms coarse-grained counterparts and horizontal vector patterns, achieving significantly higher recall at comparable sparsity levels and closely approaching the oracle statistics. This finding highlights that the vertical-vector pattern effectively captures the intrinsic fine-grained sparsity structure of attention maps. We provide extended results on statistical comparison of different sparse patterns in \Cref{sec:statComparison}. Notably, we consistently observe that the vertical-vector sparse pattern emerges across diverse modalities, closely aligning with the oracle pattern, as discussed in \Cref{sec:modality}.
\section{\name{}}
\begin{figure*}[t]
    \centering
    \includegraphics[width=0.9\linewidth]{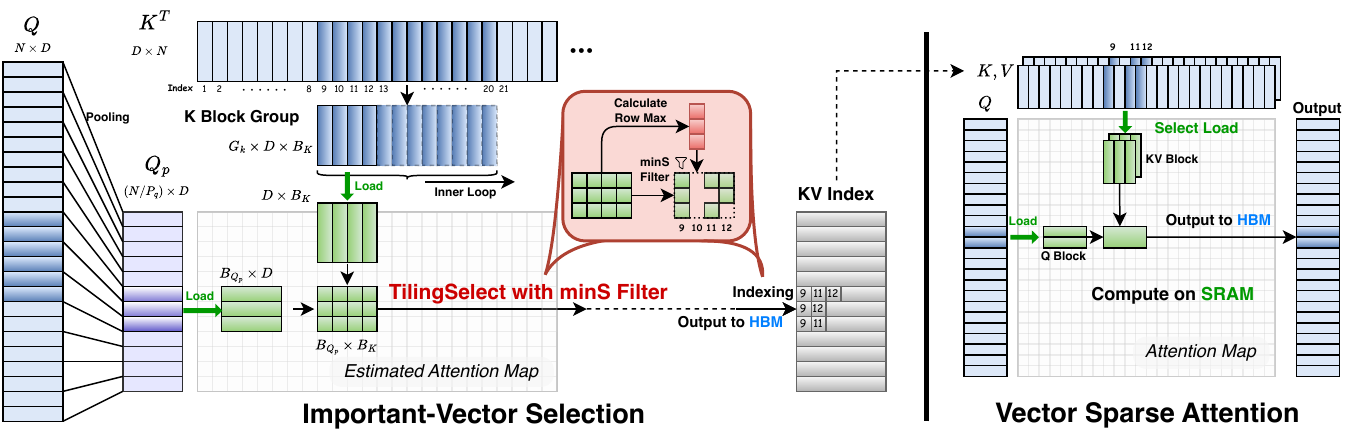}
    \caption{The overview of \name{}, which can be divided into two stages: (1) Important-Vector Selection via TilingSelect and minS Filter; (2) Vector Sparse Attention on selected vectors.}
    \label{fig:overview}
\end{figure*}

In this section, we present \name{}, a novel framework of vector-wise sparse attention that achieves significant acceleration for long-context inference. 
As illustrated in \Cref{fig:overview}, \name{} operates through two stages:
\begin{enumerate}
    \item \textbf{Important-Vector Selection:} We utilize tile-based vector selection \textit{TilingSelect} with \textit{minS} filter to select important vertical vectors with minimal HBM access. 
    \item \textbf{Vector Sparse Computation:} Attention is computed on only selected vectors using optimized GPU kernels.
\end{enumerate}
The detailed implementations of the kernel for both stages are provided in \Cref{sec:kernel}.

\subsection{Important-Vector Selection}
\label{sec:imp_vec_select}
\subsubsection{Challenges}
\label{subsec:challenges}
\textbf{Vector Importance Definition.}
In vision tasks, adjacent tokens typically correspond to pixels that are close in time or space, and thus tend to exhibit similar attention preferences toward the same key tokens~\cite{tan2025dsv}.  
Motivated by this property, we introduce a \textit{query pooling} strategy that efficiently estimates the importance of vertical vectors while substantially reducing computational overhead.

Formally, given the query matrix $\boldsymbol{Q} \in \mathbb{R}^{N \times D}$, 
we obtain the pooled queries $\boldsymbol{Q}_p \in \mathbb{R}^{N_p \times D}$ by
\begin{equation}
\label{eq:query_pooling}
\boldsymbol{Q}_p[i] = \frac{1}{P_q}\sum_{t=(i-1)P_q+1}^{iP_q} \boldsymbol{Q}[t], \quad i = 1, \ldots, N_p,
\end{equation}
where $P_q$ is a hyperparameter controlling the \textit{vector size}, 
$N$ is the input context length, and $N_p = N / P_q$ is the number of pooled query blocks. 

Each pooled query $\boldsymbol{Q}_p[i]$ serves as a compact representation of its corresponding 
query group in $\boldsymbol{Q}$, capturing their shared semantic information. Using the pooled queries $\boldsymbol{Q}_p$ and the original keys $\boldsymbol{K}$, we compute the \textit{estimated attention map} $\boldsymbol{A}_p \in \mathbb{R}^{N_p \times N}$ following 
\Cref{eq:attention_map}. 
Each element $\boldsymbol{A}_p[i,j]$ in $\boldsymbol{A}_p$ represents the estimated importance of a 
vertical vector $\boldsymbol{A}[iP_q:(i+1)P_q,j]$ in the full attention map 
$\boldsymbol{A} \in \mathbb{R}^{N \times N}$.

\textbf{The Performance Bottleneck in Fine-Grained Vector Selection.}
With the definition of vector importance, a naive approach would apply row-wise topK or topP filtering over the entire estimated attention map $\boldsymbol{A}_p$ to select the important vertical vectors. 
However, this approach introduces severe performance bottlenecks for two challenges.
\begin{itemize}
    \item \textbf{Challenge-1}: Row-wise topK/topP relies on expensive sorting operations to identify important entries, resulting in substantial computational overhead. As shown in \Cref{fig:estimateBreakdown}, the topP part alone accounts for more than 75\% of the total runtime. To address this challenge, we propose a sorting-free \textit{minS} filter in \Cref{subsec:mins}.
    \item \textbf{Challenge-2}: The naive approach must materialize the $\Theta(N^2 P_q^{-1})$ estimated attention map, leading to heavy HBM read/write traffic that dominates runtime in long-context settings. For instance, at a context length of 64K, the naive approach incurs up to 26.3 GB of HBM access (see \Cref{fig:estimateBreakdown}), resulting in heavy memory bandwidth pressure. To address this challenge, we propose \textit{TilingSelect} to fuse vector selection into the tiled-based GEMM operation in \Cref{subsec:tile}.
\end{itemize}

\subsubsection{Solution to Challenge-1: Sorting-Free minS}
\label{subsec:mins}
Existing approaches typically adopt row-wise \textit{topK}~\cite{tang2024quest,yuan2025native} or \textit{topP}~\cite{lai2025flexprefill,lin2025twilight} filtering strategies to identify important regions. 
In the \textit{topK} strategy, exactly the $k$ largest elements in each row are selected as important positions, while in the \textit{topP} (or nucleus) strategy, elements are selected such that their cumulative attention scores exceed a threshold $p$. 
However, both strategies rely on sorting operations that introduce substantial computational overhead. 
Although some implementations approximate these operations using binary search~\cite{lin2025twilight,ye2025flashinfer}, this operation remains inefficient on GPUs because its data-dependent branching and non-coalesced memory accesses cause severe warp divergence and prevent effective parallelization.

We instead propose \textit{minS} filtering, a simple sorting-free threshold-based approach that operates as follows.
For each row $i$, we first compute its maximum attention score
\[
m_i^{s} = \operatorname{rowmax}(\boldsymbol{s}_i) = \max_{j=1}^{N} \frac{\langle \boldsymbol{Q}_p[i], \boldsymbol{k}_j \rangle}{\sqrt{D}}
\]
and then apply the following threshold-based filtering rule:
\begin{equation}
\label{eqn:idx}
\boldsymbol{M}_i = (\boldsymbol{s}_i \ge  (m_i^{s}-\alpha)), \quad 
\Idx(i) = \{j \mid \boldsymbol{M}_i(j) = 1\},
\end{equation}
where $\boldsymbol{M}_i \in \{0,1\}^N$ denotes the binary mask obtained from filtering the score vector $\boldsymbol{s}_i$, 
$\alpha$ is a filtering ratio that controls sparsity, 
and $\Idx(i)$ indicates the indices of the important elements in the $i$-th row of the estimated attention map.
Beyond being sorting-free, minS further improves efficiency by applying filtering directly on the dot-product attention scores, thus avoiding the need for softmax computation and reducing memory access.

\begin{figure}[t]
    \centering
    \includegraphics[width=\linewidth]{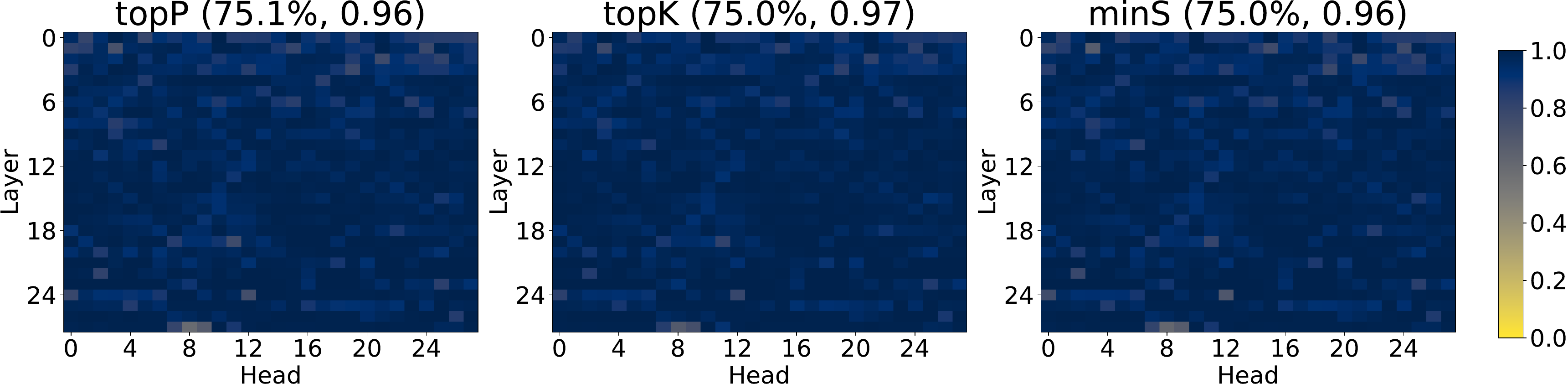}
    \caption{Recall heatmaps of different filter strategies under the same sparsity. The minS filter strategy achieves comparable recall under the same sparsity level. Markers such as (75.0\%, 0.96) denote (sparsity, recall), with sparsity in percentage.}
    \label{fig:acc_minS}
\end{figure}

As shown in \Cref{fig:acc_minS}, our minS filtering strategy achieves recall comparable to topP and topK under the same sparsity level. 
However, as illustrated in \Cref{fig:estimateBreakdown}, minS substantially reduces selection latency (3.77$\times$ speedup) due to its sorting-free and memory-access-efficient design. 
This result demonstrates that minS provides a better trade-off among recall, sparsity, and runtime efficiency in fine-grained vector selection compared to traditional topK and topP strategies.

\textbf{Dynamic Programming for Offline Search of Per-Head Filter Ratios.}
Previous studies~\cite{xu2025xattention,cai2024pyramidkv} have shown that the sparsity and importance of different attention heads vary significantly. 
Consequently, applying a uniform sparsity across all attention heads often results in suboptimal overall performance. 
To address this issue, we propose a dynamic-programming-based \textit{Per-head Filter Ratio Prediction} technique, which determines the optimal filter ratio $\alpha$ for each attention head individually to achieve the best performance under a given target sparsity constraint.

Considering a model with a total of $H$ attention heads, we define a dynamic programming table $\mathrm{DP}[h][\rho]$, where $h\in \{1, 2, \ldots, H\}$ and $\rho \in [0, 1]$.
Each entry $\mathrm{DP}[h][\rho]$ represents the optimal performance achievable by the first $h$ attention heads under an average sparsity of $\rho$.
Note that $\mathrm{DP}[0][\cdot]=0$.
The DP recurrence relation is given by
\begin{align}
\label{eqn:dp}
&\mathrm{DP}[h][\rho] \nonumber\\
&=
\max_{\alpha \geq 0} \!\biggl\{
    \mathrm{DP}\bigl[h-1\bigr]\!\bigl[\frac{\rho\!\cdot\!h-\mathrm{sp}_h(\alpha)}{h-1}\bigr]\!\!
+\! \operatorname{Perf}_h(\alpha)
\biggr\},
\end{align}
where $\mathrm{sp}_h(\alpha)$ denotes the sparsity of the $h$-th attention head corresponding to a given filter ratio $\alpha$, 
and $\operatorname{Perf}_h(\alpha)$ measures the corresponding performance of the $h$-th head.
The functions $\mathrm{sp}_h(\alpha)$ and $\operatorname{Perf}_h(\alpha)$ can be recorded offline by sampling a set of candidates.
Under the target sparsity $\rho_T$, the optimal overall performance is given by $\mathrm{DP}[H][\rho_T]$. 

\subsubsection{Solution to Challenge-2: Tile-Fused TilingSelect} 
\label{subsec:tile}

To alleviate the memory traffic brought by materializing large estimated attention map, we propose \textit{TilingSelect}, which fuses the selection operation directly into the tiled GEMM computation of $\boldsymbol{S}_p = \boldsymbol{Q}_p \boldsymbol{K}^\top / \sqrt{D}$. 

\begin{figure}[t]
    \centering
    \includegraphics[width=0.9\linewidth]{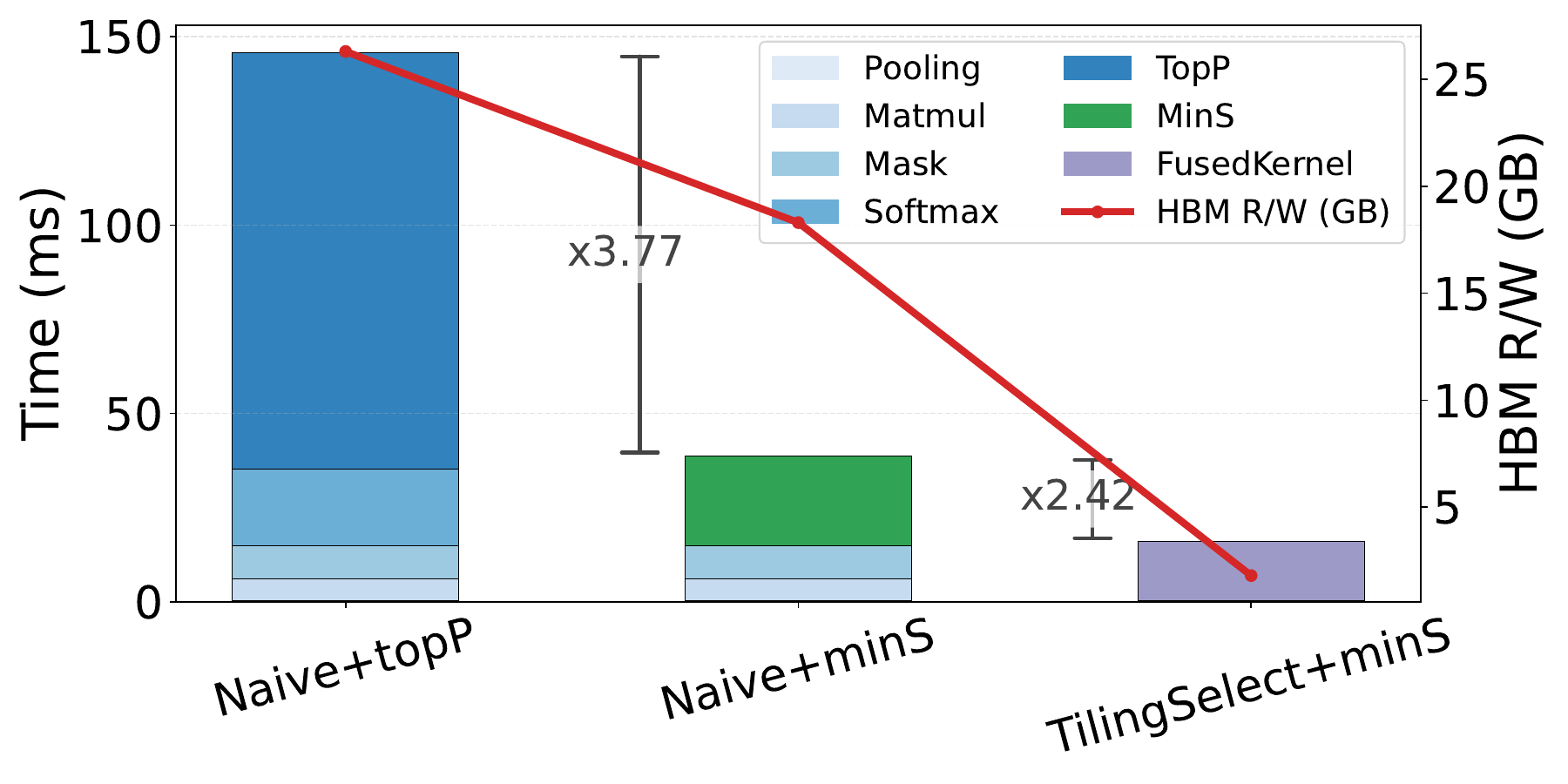}
    \caption{
Time and memory-traffic breakdown at context length $N{=}64\mathrm{K}$ and sparsity $0.9$.
Compared with a naive materialize-then-filter baseline (e.g., topP), \emph{TilingSelect} with \emph{minS} avoids writing estimated attention map to HBM, cutting memory traffic from 18.3\,GB to 1.8\,GB and reducing selection latency by 2.42$\times$, yielding an overall 9.12$\times$ speedup.
    }
    \label{fig:estimateBreakdown}
\end{figure}

Specifically, each query tile $\boldsymbol{Q}_p^{\text{tile}} \in \mathbb{R}^{B_{Q_p} \times D}$ and key tile 
$\boldsymbol{K}^{\text{tile}} \in \mathbb{R}^{B_K \times D}$ are loaded into on-chip memory to compute 
$\boldsymbol{S}_p^{\text{tile}} = \boldsymbol{Q}_p^{\text{tile}} (\boldsymbol{K}^{\text{tile}})^\top / \sqrt{D} \in \mathbb{R}^{B_{Q_p} \times B_K}$. 
Instead of writing $\boldsymbol{S}_p^{\text{tile}}$ back to HBM, we immediately apply the minS filtering strategy and retain only the indices of important elements. 
This fused design reduces HBM access from $\Theta(N^2 P_q^{-1})$ to $\Theta(N^2 P_q^{-1}(1-\rho))$, where $\rho$ denotes the sparsity level. 
Under high sparsity settings (see \Cref{fig:sparsePattern,fig:visualizationMap}), TilingSelect significantly lowers HBM traffic and thus reduces latency. As confirmed in \Cref{fig:estimateBreakdown}, at a sparsity level of 0.9, TilingSelect with minS filtering requires only 1.8 GB of memory access—far lower than 18.3 GB required by the naive minS approach—yielding a 2.42$\times$ speedup in vector selection.

\textbf{System-Algorithm Co-optimization.}
During the tiling-based selection, we introduce an additional parameter $G_k$, 
which denotes the number of $K$-tiles of shape $B_K \times D$ processed by each thread block. 
This optimization brings two major benefits.
(1) \textit{Controlling thread-block parallelism.}  
When the input length $N$ is small or the vector size $P_q$ is large, 
the pooled query matrix $\boldsymbol{Q}_p$ becomes relatively short, 
resulting in a flat estimated attention map. 
In this case, relying solely on $\boldsymbol{Q}_p$-dimension parallelism 
leads to under-utilization of GPU computational resources during the flat GEMM operation. 
By decreasing $G_k$, we enable higher parallelism along the $K$ dimension, 
thereby improving SM occupancy and achieving better hardware utilization.
(2) \textit{Cross-tile minS optimization.}  
When a thread block iteratively processes multiple $K$-tiles, 
we maintain a per-row running maximum (\textit{row-max accumulator}) across tiles. 
This row-max accumulator enables a more global thresholding in the subsequent tiles, 
where each tile can reuse the accumulated maximum to perform cross-tile \textit{minS} filtering. 
This accumulation mechanism enhances filtering consistency, leading to improved accuracy.

\subsection{Vector-Sparse Computation}
\label{sec:vector_sparse_comp}

After selecting important vectors, 
we perform attention computation on only these vectors. 
Unlike conventional block-sparse implementations that load contiguous key--value tiles, 
our \textit{vector-sparse attention} dynamically fetches discrete keys and values with the selected key-value indices, 
following the sharding strategy of FlashAttention~\cite{dao2023flashattention}.

Specifically, the set $\Idx(i)$ from \Cref{eqn:idx}, which represents the positions of vertical vectors in attention map for $i$-th query block (see \Cref{eq:query_pooling}), contains the indices of the important key-value vectors to be loaded. 
We then compute the sparse attention output for this query block as
\begin{equation}
\label{eq:vector_sparse_out}
\begin{split}
\boldsymbol{O}[\mathcal{I}_B(i)] &=
\operatorname{softmax}\!\left(
\frac{
\boldsymbol{Q}[\mathcal{I}_B(i)]\,\boldsymbol{K}[\Idx(i)]^\top
}{
\sqrt{D}
}
\right)
\boldsymbol{V}[\Idx(i)],
\end{split}
\end{equation}
where the slicing operator $[\cdot]$ extracts the corresponding 
rows from the full $\boldsymbol{Q}$, $\boldsymbol{K}$, or $\boldsymbol{V}$ matrices. $\mathcal{I}_B(i) = \{j|j\in [iP_q:(i+1)P_q]\}$ 
represents the row indices of the $i$-th query block (the query block size is set equal to the vector size $P_q$), 
and $\Idx(i)$ is the index set selecting a subset of 
$\boldsymbol{K}[\Idx(i)] \in \mathbb{R}^{n_i \times D}$ 
and $\boldsymbol{V}[\Idx(i)] \in \mathbb{R}^{n_i \times D}$, 
with $n_i$ denoting the number of selected key-value vectors. 

During kernel execution, only the selected key-value vectors $\boldsymbol{K}[\Idx(i)]$ and $\boldsymbol{V}[\Idx(i)]$ are loaded into on-chip memory, and all intermediate attention scores and outputs are accumulated locally. This vector-sparse computation minimizes HBM traffic and preserves high arithmetic throughput, while effectively leveraging the vector-sparse patterns of video attention.

\section{Evaluation}
\subsection{Setup}
\textbf{Models.}
We evaluate \name{} on two categories of video tasks: \textit{video understanding} and \textit{video generation}.
For video understanding, we employ two VLMs with distinct architectures: \textit{Qwen2.5-VL-Instruct}~\cite{qwen2.5-VL} and \textit{InternVL-3.5-8B}~\cite{wang2025internvl3_5}.
For video generation, we adopt two state-of-the-art DiTs: \textit{Wan2.1-T2V-14B}~\cite{wan2025wan} and \textit{HunyuanVideo-T2V-13B}~\cite{kong2024hunyuanvideo}.
Together, these models represent the current frontier in both video understanding and video generation.

\textbf{Benchmarks.}
We evaluate \name{} on a diverse suite of video benchmarks. 
For video understanding, we use three representative benchmarks—\textit{VideoMME}~\cite{fu2025video}, \textit{LongVideoBench}~\cite{wu2024longvideobench}, and \textit{VCRBench}~\cite{qi2025vcr}—which span a broad range of task types including perception, recognition, OCR, and high-level reasoning. 
For video generation, we employ \textit{VBench}~\cite{huang2024vbench} to assess the fidelity and consistency of the generated videos. 

\textbf{Baselines.}
We compare \name{} against strong baselines on tasks of both video understanding and video generation. 
For video understanding, we benchmark against \textit{FlexPrefill}~\cite{lai2025flexprefill}, \textit{XAttention}~\cite{xu2025xattention}, and \textit{AnchorAttention}~\cite{zhang2025anchorattention}, which is a recent method based on \textit{stripe granularity}. Notably, AnchorAttention employs a stripe-based granularity conceptually similar to our vertical-vector granularity, but with substantially larger stripe sizes that result in coarser sparse patterns.%

To ensure fair comparison, we meticulously follow the configurations from original papers and official implementations. 
For FlexPrefill, we use $\tau = 0.1$; for AnchorAttention, step $= 16$ and block size $= 128$; for XAttention, $S = 16$ in video understanding tasks. 
We systematically sweep each method's sparsity-controlling parameters ($\gamma$ for FlexPrefill, $\tau$ for XAttention, $\theta$ for AnchorAttention) 
to align sparsity levels before accuracy comparison. For video generation, we compare against XAttention ($S = 8$) and Sparse VideoGen~\cite{xi2025sparse}. The Full Attention baseline is implemented using \textit{FlashAttention-2}~\cite{dao2023flashattention}. 
For \name{}, we set the vector size $P_q = 64$, key-block size $B_k = 16$, and key-block group size $G_k = 16$ for video understanding and $G_k = 8192$ for video generation (due to consistently long sequences, where high Q-dimension parallelism suffices). The filtering ratio $\alpha$ is dynamically adjusted to control sparsity. All experiments are conducted on NVIDIA A100 (80GB) GPUs. 

\subsection{Accuracy Results}

\begin{table}[t]
    \centering
    \small
    \caption{Performance of different models and different methods on video understanding tasks. All values are percentages (i.e., multiplied by 100) without the \% symbol.}
    \label{tab:video_understanding}
    \begin{tabular}{l r r r r r}
        \toprule
        {\textbf{Model}} & 
        {\parbox{0.7cm}{\centering \textbf{Avg.} \\ \textbf{Sp.}}} & 
        {\parbox{0.8cm}{\centering \textbf{Video} \\ \textbf{MME}}} &
        {\parbox{0.8cm}{\centering \textbf{Long} \\ \textbf{Video}}} & 
         {\parbox{0.8cm}{\centering \textbf{VCR-} \\ \textbf{Bench}}} & 
        {\parbox{0.8cm}{\centering \textbf{Avg.} \\ \textbf{Acc.}}} \\
        \midrule
        \multicolumn{6}{c}{\textit{InternVL-3.5-8B \quad \#Frames: 64 \#Tokens: 17K}} \\
        \midrule
        \rowcolor{Gray}
        Full Atten. & 0.0 & 65.7 & 59.4 & 32.9 & 52.7 \\
        FlexPref. & 76.5 & 52.3 & 59.0 & 30.0 & 47.1 \\
        XAtten. & 78.1 & 56.0 & \textbf{59.9} & 32.5 & 49.5 \\
        AnchorAtten. & 78.6 & 57.4 & 59.4 & 31.3 & 49.4 \\
        \rowcolor{lightblue}
        \textbf{VecAtten.} & \textbf{78.6} & \textbf{60.6} & 59.0 & \textbf{33.8} & \textbf{51.1} \\
        \midrule
        \multicolumn{6}{c}{\textit{Qwen2.5-VL-7B-Instruct \quad \#FPS: 1 \#Tokens: 26K}} \\
        \midrule
        \rowcolor{Gray}
        Full Atten. & 0.0 & 63.9 & 59.9 & 25.8 & 49.9 \\
        FlexPref. & 73.6 & 62.0 & 56.7 & 22.5 & 47.1 \\
        XAtten. & 73.6 & 63.0 & 58.5 & 20.0 & 47.2 \\
        AnchorAtten. & 74.6 & 64.4 & \textbf{60.8} & 22.9 & 49.4 \\
        \rowcolor{lightblue}
        \textbf{VecAtten.} & \textbf{78.5} & \textbf{64.8} & 59.4 & \textbf{25.4} & \textbf{49.9} \\
        \bottomrule
    \end{tabular}
\end{table}

\textbf{Video Understanding.}
As evidenced by \Cref{tab:video_understanding}, \name{} consistently achieves superior accuracy-sparsity trade-offs across almost all video understanding benchmarks. 
Notably, \name{} maintains near-full accuracy while operating at significantly higher sparsity levels compared to baselines. 
For the Qwen2.5-VL-7B-Instruct model, \name{} matches Full Attention performance on long-video inputs (1 frame per second sampling) while achieving 78.5\% average sparsity—a remarkable compression that demonstrates the effectiveness of our vector-wise sparse pattern. The advantage is particularly pronounced on complex reasoning tasks (VCRBench) where precise attention to specific temporal regions is crucial. This result aligns with our core insight that fine-grained vertical-vector sparsity better preserves critical information compared to coarse-grained alternatives.

\begin{table}[t]
\centering
\caption{Performance of different models and methods on video generation tasks, with 50 inference steps and a 6\% warm-up ratio.}
\label{tab:video_generation}
\renewcommand{\arraystretch}{1.3}
\resizebox{0.95\linewidth}{!}{%
\begin{tabular}{l r r r r}
\toprule
\textbf{Model} & \textbf{Sparsity(\%)} $\uparrow$ & \textbf{PSNR} $\uparrow$ & \textbf{SSIM} $\uparrow$ & \textbf{LPIPS} $\downarrow$ \\
\midrule
\multicolumn{5}{c}{\textit{Wan 2.1-T2V-14B \quad 720P, \#Tokens 76K} } \\
\cmidrule(lr){1-5}
XAttention      & 54.6 & 19.7 & 0.658 & 0.348 \\
SVG             &  52.2& 18.7 & 0.639 & 0.381\\ 
\rowcolor{lightblue}
\name{}         & 52.3 & \textbf{19.7} & \textbf{0.668} &\textbf{0.339} \\ 
\midrule
\multicolumn{5}{c}{\textit{HunyuanVideo-T2V-13B \quad 720P, \#Tokens 119K}} \\
\cmidrule(lr){1-5}
XAttention      & 60.8 & 21.2 & 0.734 & 0.348 \\ 
SVG             &  60.1& 21.8 & 0.769 & \textbf{0.326}\\ %
\rowcolor{lightblue}
\name{}          & \textbf{62.1}& \textbf{22.8} & \textbf{0.779} &  0.330 \\
\bottomrule
\end{tabular}
}
\end{table}

\textbf{Video Generation.}
\name{} is applied to Wan2.1 and HunyuanVideo DiT models to generate long videos 
(average context lengths: 76K and 119K tokens, respectively) using VBench prompts. 
As shown in \Cref{tab:video_generation}, under comparable sparsity levels, \name{} consistently outperforms other sparse-attention baselines across all quantitative metrics (PSNR, SSIM, and LPIPS). Qualitative comparisons in \Cref{sec:qualitative} provide compelling visual evidence: for identical prompts, videos generated by \name{} exhibit superior visual fidelity and temporal consistency, closely matching Full Attention outputs. In contrast, baseline methods introduce noticeable visual artifacts and inconsistent scene elements, particularly in complex scenes requiring long-range temporal dependencies.

\subsection{Latency Results}
\begin{figure*}[t]
    \centering
    \begin{subfigure}[b]{0.3\textwidth}
        \centering
        \includegraphics[width=\textwidth]{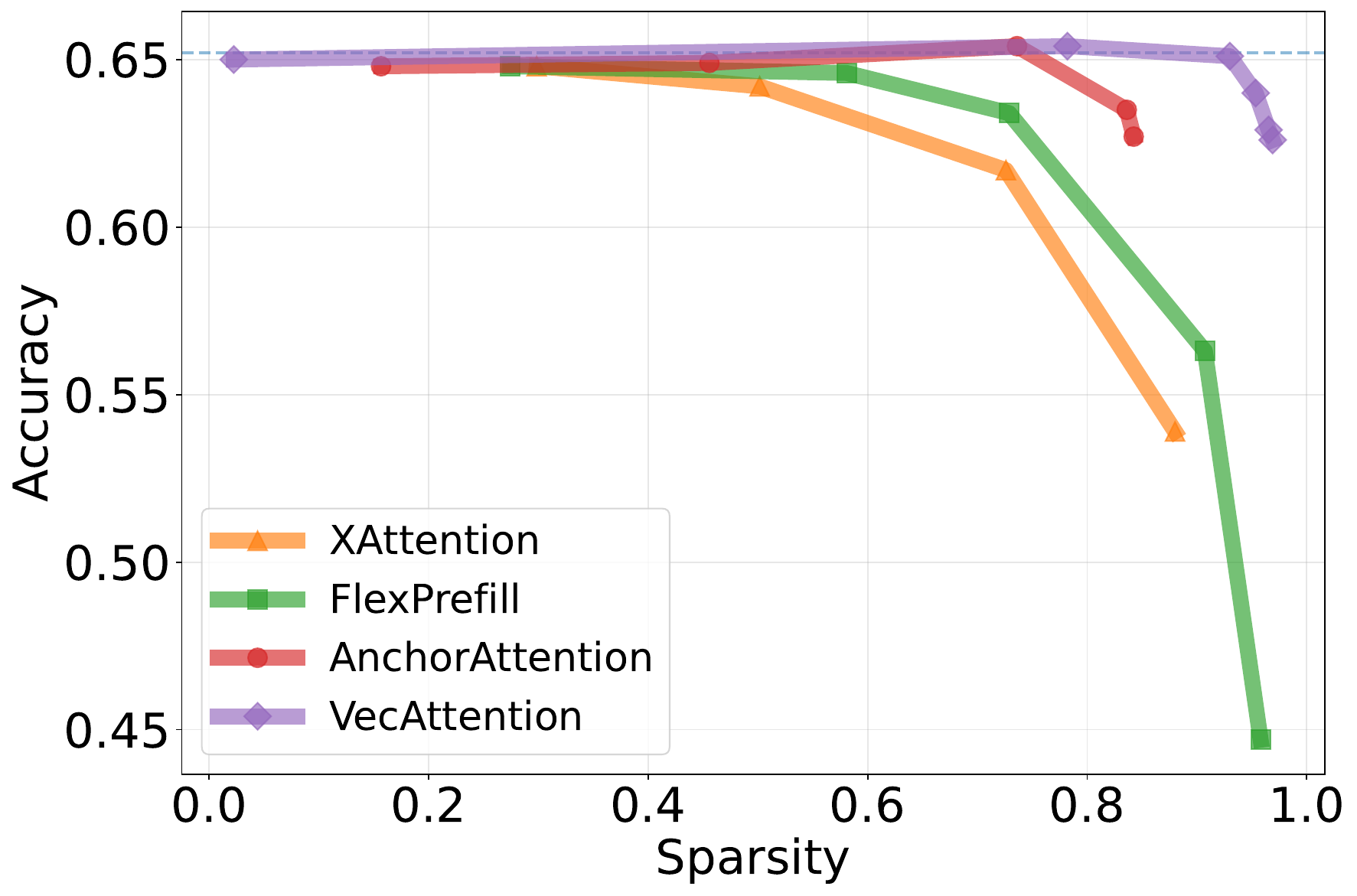}
        \caption{Sparsity vs. Accuracy}
    \end{subfigure}
    \hfill
    \begin{subfigure}[b]{0.3\textwidth}
        \centering
        \includegraphics[width=\textwidth]{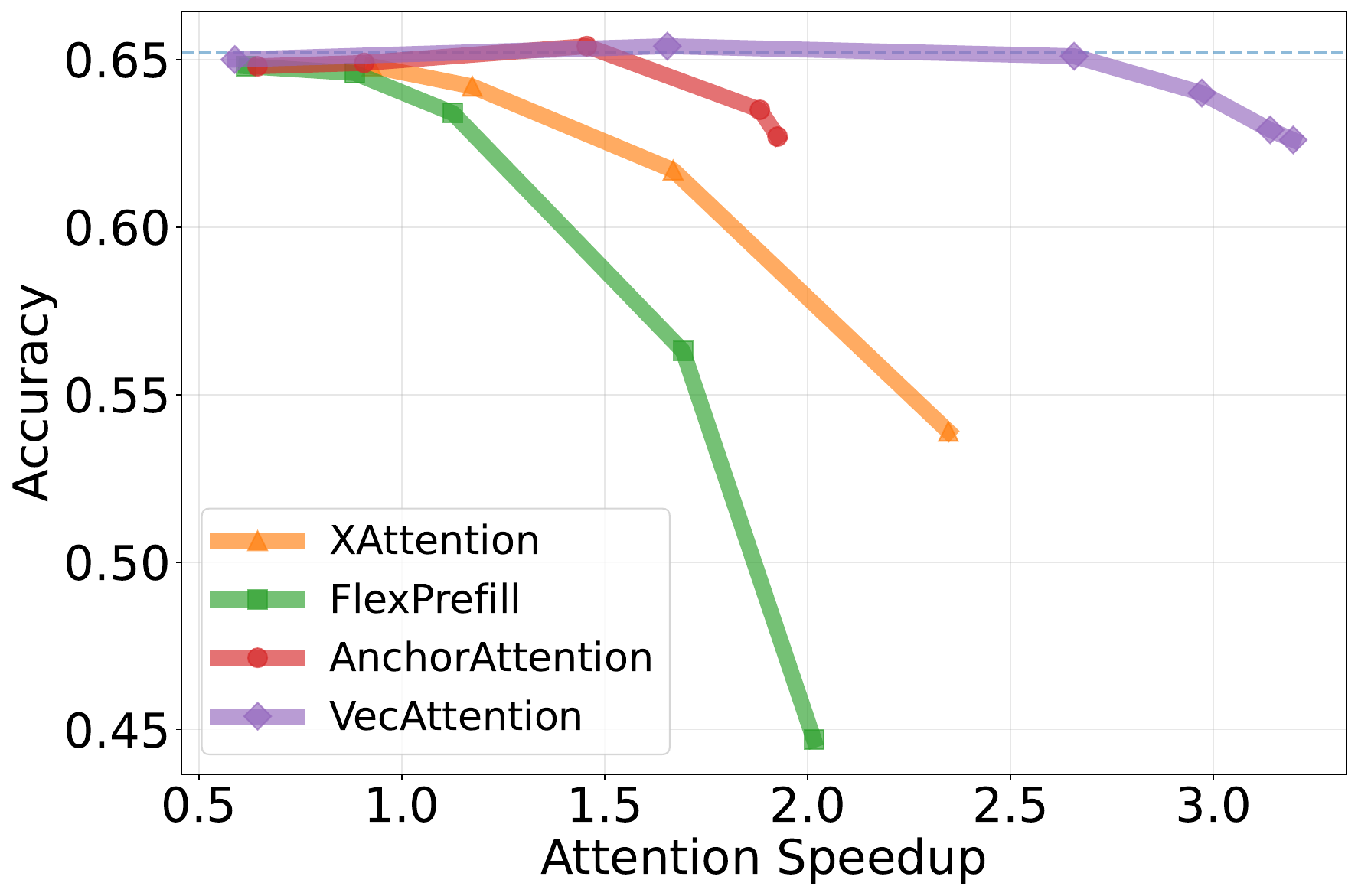}
        \caption{Attention Speedup vs. Accuracy}
    \end{subfigure}
    \hfill
    \begin{subfigure}[b]{0.3\textwidth}
        \centering
        \includegraphics[width=\textwidth]{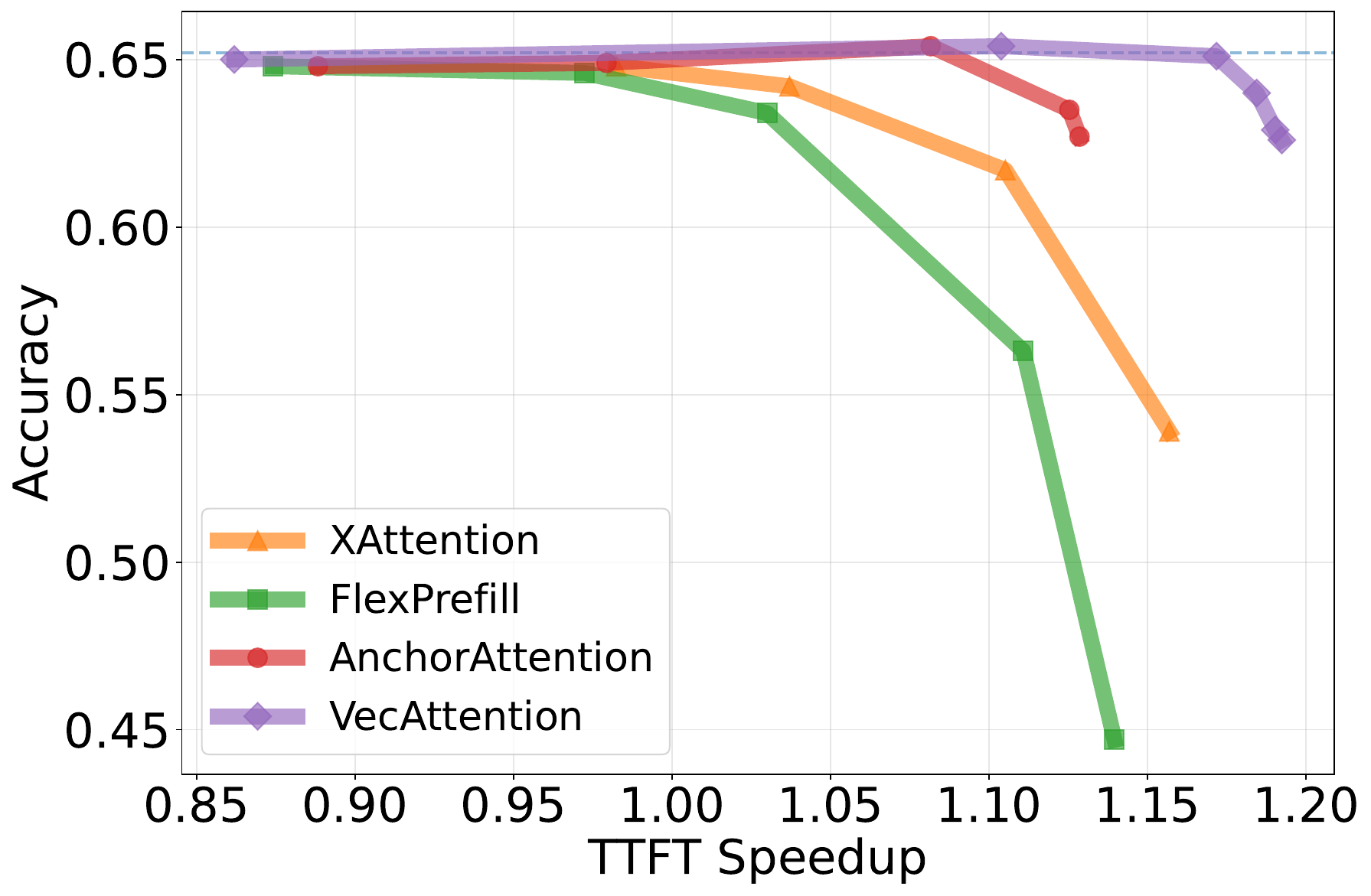}
        \caption{TTFT Speedup vs. Accuracy}
    \end{subfigure}
    \caption{
Microbenchmarks of efficiency-accuracy trade-offs across different sparse attention methods on VideoMME. \name{} preserves accuracy at up to 93\% sparsity and translates it into practical gains, which are 2.65$\times$ acceleration for attention and 1.17$\times$ end-to-end TTFT.
}
    \label{fig:microbenchmark}
\end{figure*}

\textbf{Microbenchmarks.}
We conduct detailed microbenchmarks on VideoMME to quantify the accuracy-efficiency trade-offs of \name{}. 
By sweeping sparsity thresholds across all methods, we establish comprehensive Pareto frontiers. As demonstrated in \Cref{fig:microbenchmark}, our fine-grained vertical-vector sparse pattern enables \name{} to reach remarkable 93\% sparsity without compromising accuracy, substantially exceeding the coarse-grained patterns.

The optimized techniques of important-vector selection in \name{} translate this theoretical advantage into practical speedups, yielding up to \(2.65\times\) acceleration for attention computation and \(1.17\times\) improvement in end-to-end Time-To-First-Token (TTFT) speedup. This result demonstrates that our approach not only reduces FLOPs but also efficiently handles the system-level challenges of fine-grained sparsity.

\textbf{Attention Computation Breakdown.}
\Cref{fig:sparsePattern:c} further breaks down the runtime contributions of important-region selection and sparse attention computation under accuracy-preserving settings. 
We can see that the superior speedup of \name{} primarily comes from two factors: 
(1) the finer-grained sparsity enables substantially higher overall sparsity levels, reducing the cost of sparse attention computation; and 
(2) the optimized techniques in important-vector selection keep the overhead in line with coarse-grained baselines, or even lower. This breakdown analysis validates that \name{} avoids the common pitfall where selection overhead negates the benefits of increased sparsity.

\subsection{Ablation Study}
\begin{figure}[t]
    \centering
    \begin{subfigure}[b]{0.45\linewidth}
        \centering
        \includegraphics[width=\textwidth]{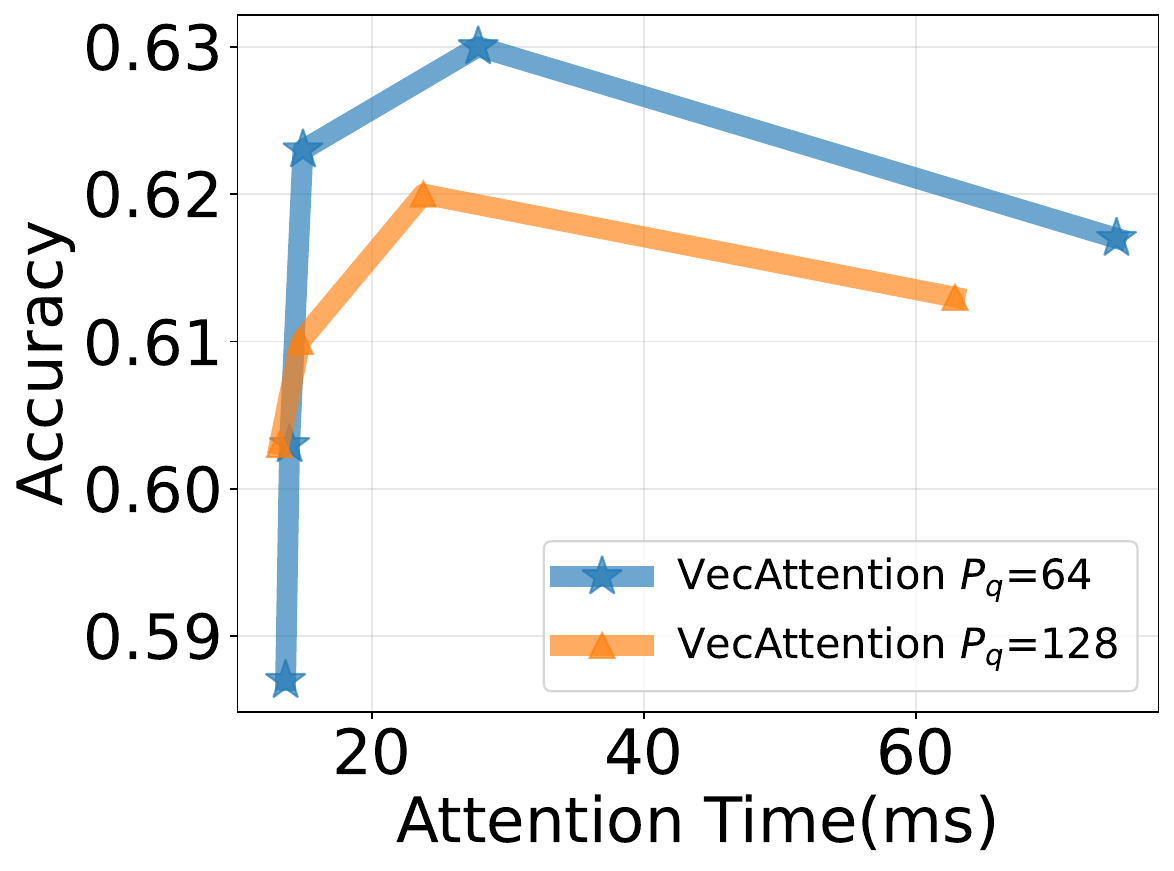}
        \caption{Effects of Vector Size $P_q$}
        \label{fig:abl_Pq}
    \end{subfigure}
    \hfill
    \begin{subfigure}[b]{0.45\linewidth}
        \centering
        \includegraphics[width=\textwidth]{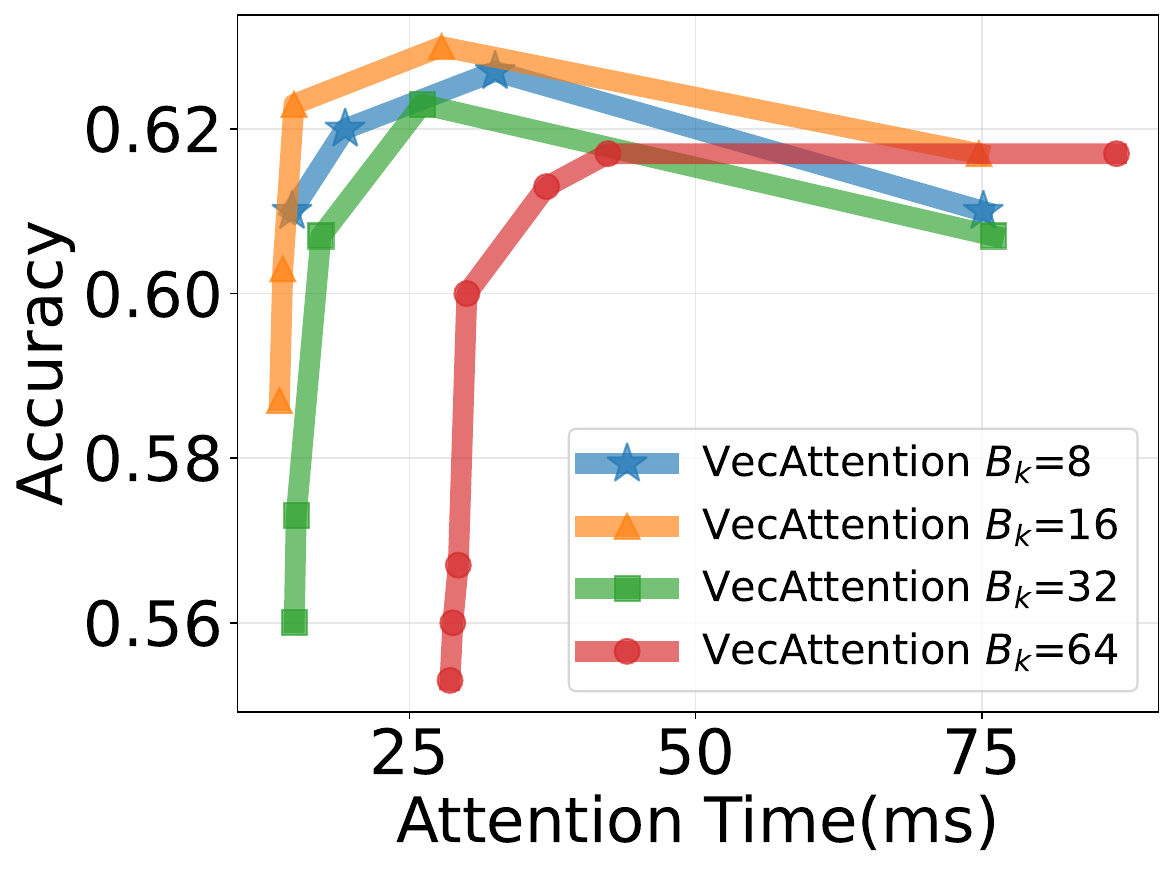}
        \caption{Effects of K Tile Size $B_k$}
        \label{fig:abl_Bk}
    \end{subfigure}
    \\
    \begin{subfigure}[b]{0.5\linewidth}
        \centering
        \includegraphics[width=\textwidth]{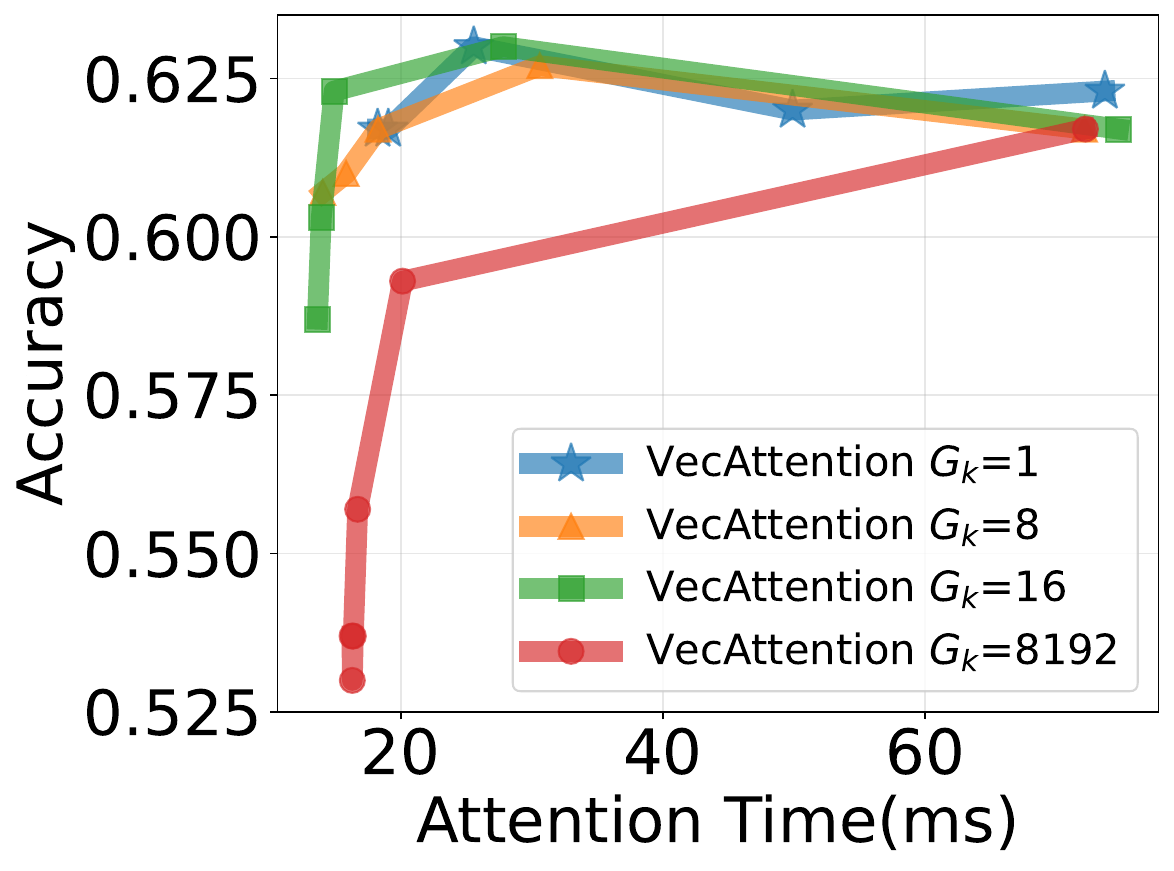}
        \caption{Effects of K-Block Group Size $G_k$}
        \label{fig:abl_Gk}
    \end{subfigure}
    \caption{Ablation study on VideoMME, which contains diverse video inputs ranging from 11 seconds to 1 hour in duration, for three key parameters.}
    \label{fig:ablation}
\end{figure}

As shown in \Cref{fig:ablation}, each parameter plays a distinct role in the accuracy-efficiency trade-off.

\begin{itemize}
\item \textbf{Vector Size ($P_q$)} governs the granularity of our sparse pattern. 
Increasing $P_q$ improves latency through reduced estimation cost but eventually degrades accuracy due to coarser representation. 
The optimal value $P_q=64$ balances these competing objectives, providing sufficient granularity for accurate selection 
while maintaining efficient implementation.

\item \textbf{K Tile Size ($B_k$)} controls memory-compute balance within thread blocks. 
Larger $B_k$ values increase shared memory and register usage, reducing thread-block concurrency but enabling richer local candidate sets for selection. 
Our chosen $B_k=16$ maximizes SM occupancy while maintaining selection quality.

\item \textbf{Group Size ($G_k$)} enables cross-tile optimization by controlling serial iteration depth over K-tiles. 
While larger $G_k$ prolongs register lifetimes and reduces memory-compute overlap, it facilitates cross-tile minS accumulation for globally consistent filtering. The optimal $G_k=16$ provides substantial accuracy improvements with low computational overhead.
\end{itemize}

This analysis demonstrates that \name{}'s parameters are not merely implementation details but fundamental design choices that systematically balance algorithmic effectiveness and hardware efficiency.

\section{Related Work}

\subsection{Long-Context Video Models}

We consider two major families of video models~\cite{zhou2024survey}: vision-language models (VLMs)~\cite{tang2025video,nguyen2024video,han2024free} for video understanding and diffusion transformers (DiTs)~\cite{li2023faster,croitoru2023diffusion,yang2023diffusion} for video generation. 
VLMs have evolved from image-based architectures~\cite{han2024free} with temporal adapters or token schedulers into large multimodal models capable of long-context reasoning and cross-modal grounding, as demonstrated by systems such as 
VideoLLaMA 3~\cite{zhang2025videollama}, Qwen2.5-VL~\cite{qwen2.5-VL}, and InternVideo2~\cite{wang2024internvideo2}. 
Meanwhile, DiTs have become the prevailing backbone for high-quality video synthesis, transitioning from U-Net-based diffusion models~\cite{li2023faster} to attention-centric architectures that capture long-range spatial-temporal structure, exemplified by models such as Wan2.1~\cite{wan2025wan} and HunyuanVideo~\cite{kong2024hunyuanvideo}. 
However, both VLMs and DiTs suffer from rapidly increasing attention costs as video length and resolution grow, making efficient long-context computation a central challenge.

\subsection{Sparse Attention}

The quadratic complexity of attention has motivated extensive research on {sparse attention}~\cite{sun2025efficient,ijcai24xformer}, which aims to exploit the inherent sparsity of attention maps to reduce redundant computation while preserving model accuracy.
Prior work~\cite{beltagy2020longformer,zaheer2020big,child2019generating,yuan2025native,lu2025moba,gao2024seerattention} exploits sparsity by modifying the attention architecture itself, 
typically requiring costly retraining or task-specific fine-tuning to preserve accuracy. 
Methods focus on the memory-bound {decoding stage}—e.g., token-eviction techniques~\cite{zhang2023h2o,xiao2023efficient,oren2024transformers} such as H2O~\cite{zhang2023h2o}, StreamingLLM~\cite{xiao2023efficient}, and selective-loading techniques~\cite{tang2024quest,zhang2025pqcache,liu2025clusterkv} such as Quest~\cite{tang2024quest}—whose optimizations do not transfer to the {compute-bound prefill stage}. In contrast, prefill attention and DiT denoising exhibit different sparsity and arithmetic intensity, leaving significant opportunities for acceleration that remain largely unexplored.

To accelerate prefill computation, recent methods primarily rely on coarse-grained importance estimation. 
MInference~\cite{jiang2024minference} identifies a {line pattern} in LLMs—using vertical lines and slash blocks to speed up prefill inference. 
FlexPrefill~\cite{lai2025flexprefill} builds upon this pattern and dynamically determines whether each attention head should adopt a line or block sparse pattern. 
MMInference~\cite{li2025mminference} extends line-pattern sparsity to VLMs through a feature-permutation grid strategy. 
XAttention~\cite{xu2025xattention} and SpargeAttention~\cite{zhang2025spargeattn} generalize block-sparse attention across LLMs, VLMs, and DiTs, improving the applicability of block patterns. 
In addition, many block-based sparsity methods leverage spatiotemporal redundancy in DiTs, including STA~\cite{zhang2025fast}, Radial Attention~\cite{li2025radial}, Sparse VideoGen~\cite{xi2025sparse}, and Sparse VideoGen2~\cite{yang2025sparse}. 

We also note the emergence of multiple methods that explore finer-grained sparse patterns. 
AnchorAttention~\cite{zhang2025anchorattention} identifies static {anchor positions} in LLMs—regions that consistently remain important—and derives a stripe sparse pattern by comparing token importance against these anchors. 
FG-Attention~\cite{durvasula2025fg} constructs an asynchronous gather-load mechanism in DiTs to enable more fine-grained sparse computation. 
However, neither of these methods analyzes in depth the performance overhead of important-region selection under fine-grained sparsity. We instead adopt TilingSelect with minS filtering to reduce this overhead and enable efficient vector-wise sparse attention.

\section{Conclusion}
In this paper, we have presented \name{}, a framework of vector-wise sparse attention that exploits the intrinsic vertical-vector sparsity in video attention maps to substantially reduce computational overhead while preserving model accuracy. By coupling an efficient important-vector selection strategy with an optimized kernel of vector sparse attention, \name{} achieves a more favorable accuracy-efficiency balance than previous SOTA. 
Extensive experiments on tasks of long-context video understanding and video generation demonstrate that \name{} achieves up to 2.65$\times$ speedup over full attention and 1.83$\times$ over prior sparse attention methods without degrading performance.

\section*{Acknowledgments}
We thank the anonymous reviewers for their constructive feedback. This work was partially supported by National Natural Science Foundation of China (Grants No. U25A6023, 92464301, and 62502305), Fundamental and Interdisciplinary Disciplines Breakthrough Plan of the Ministry of Education of China (No. JYB2025XDXM118), and Natural Science Foundation of Shanghai (Grant No. 25ZR1402275).
Tao Xie is also affiliated with the Key Laboratory of High Confidence Software Technologies (Peking University), Ministry of Education China; Beijing Tongming Lake Information Technology Application Innovation Center; Institute of Systems for Advanced Computing at Fudan University, Shanghai, China; Shanghai Research Institute of Systems of Open Computing, Shanghai, China.
{
    \small
    \bibliographystyle{ieeenat_fullname}
    \bibliography{main}
    \balance{}
}
\clearpage
\setcounter{page}{1}
\maketitlesupplementary

\section{Visualization of Attention Maps}
\label{sec:visualization}
\begin{figure*}[t]
    \centering
    \begin{subfigure}[b]{0.4\linewidth}
        \centering
        \includegraphics[width=\linewidth]{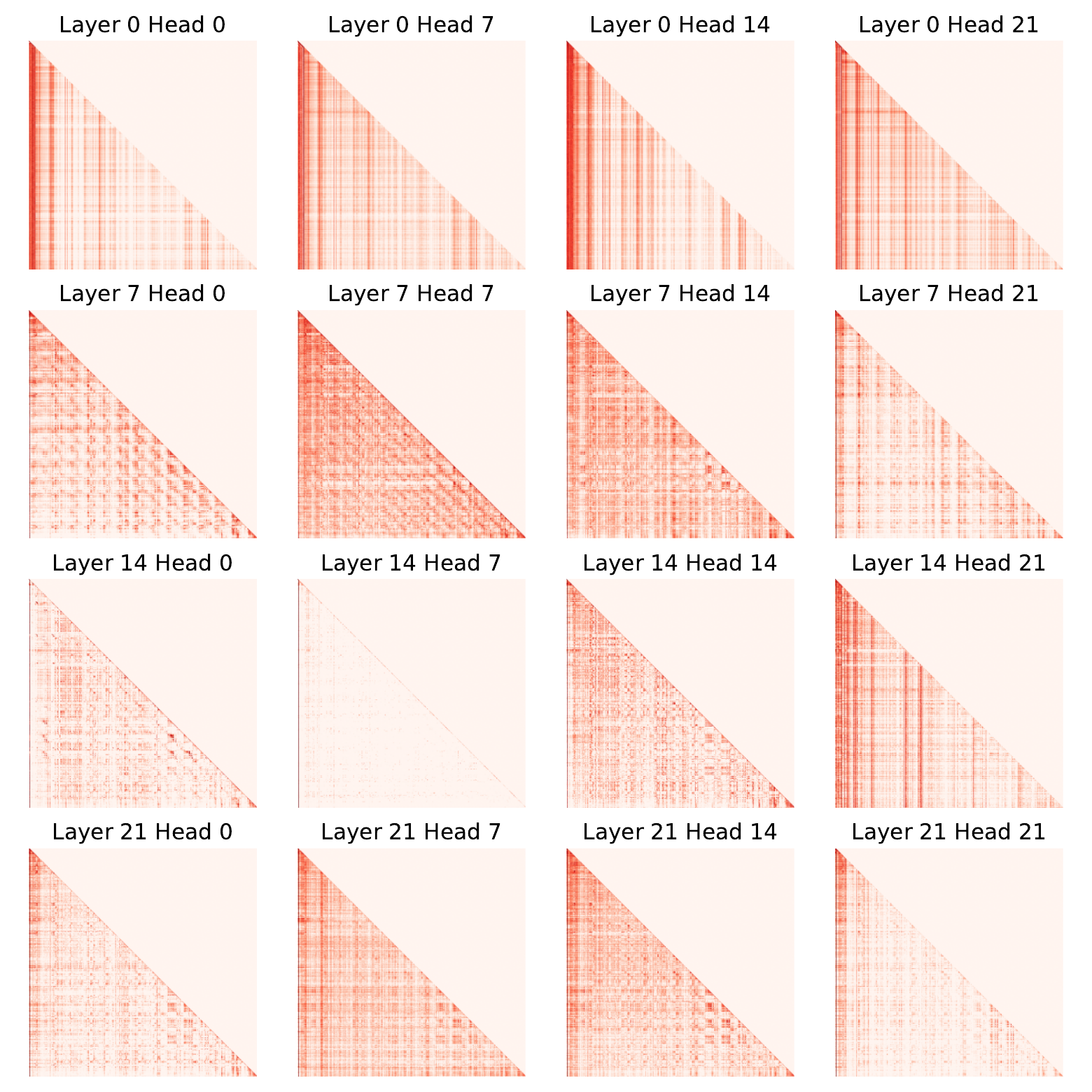}
        \caption{VideoMME (Qwen2.5-VL-7B-Instruct)}
    \end{subfigure}
    \begin{subfigure}[b]{0.4\linewidth}
        \centering
        \includegraphics[width=\linewidth]{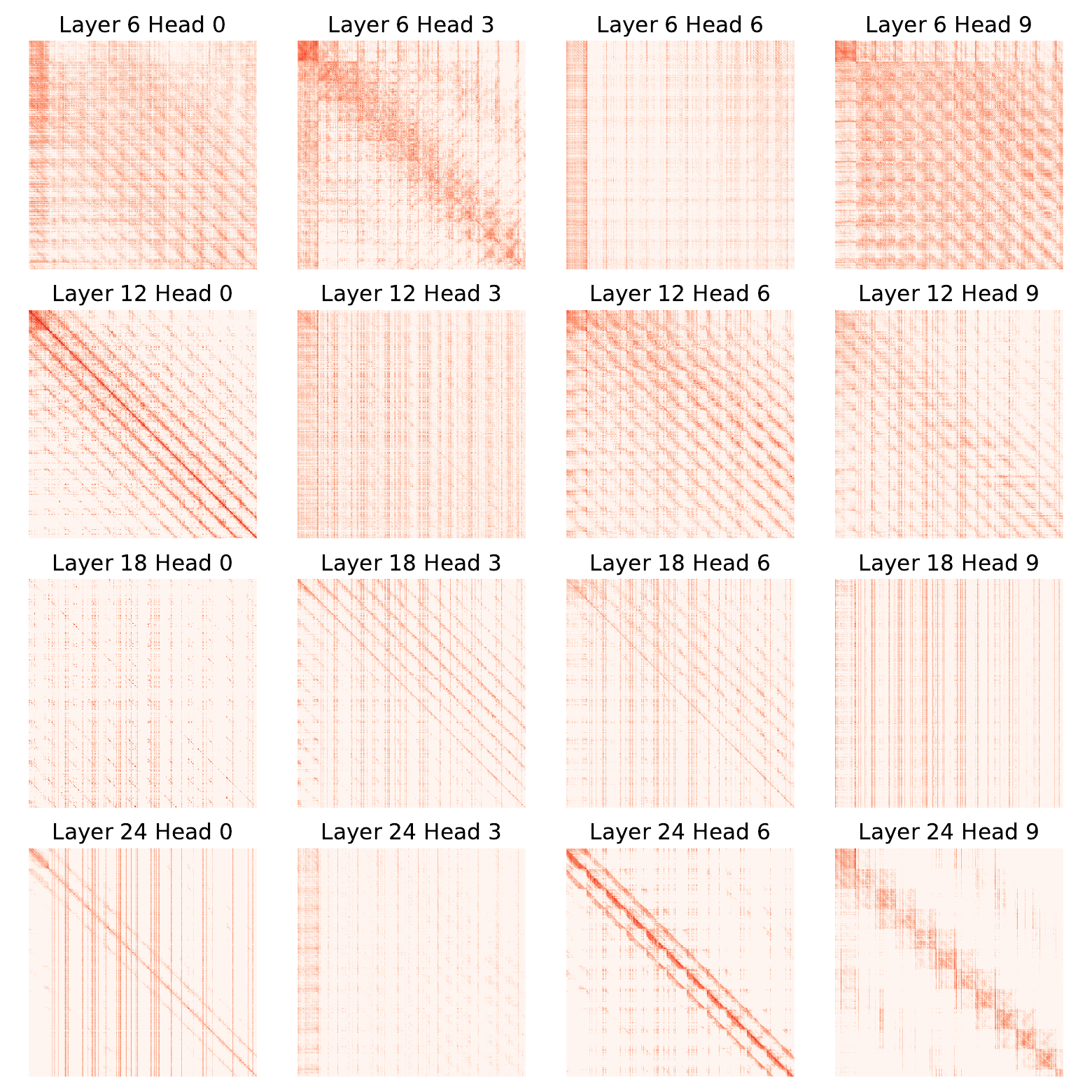}
        \caption{VBench (Wan2.1)}
    \end{subfigure}
    \caption{
   Visualization of attention maps across various layers and attention heads in tasks of video understanding and video generation.
    }
    \label{fig:morevisualizationMap}
\end{figure*}

\Cref{fig:morevisualizationMap} presents visualizations of attention maps across different layers and heads when performing inference on video benchmarks using the Qwen2.5-VL-7B-Instruct and Wan2.1 models, respectively. It can be observed that salient regions (highlighted in red) often manifest as fine-grained vertical line segments (vertical-vector pattern). It should be noted that, in theory, the vertical-vector pattern can be grouped into coarser structures such as blocks or lines. For instance, a vertical line pattern can be formed by aggregating all vertical vectors within the same column, while a diagonal (slash) line pattern can be constructed by selecting vertical vectors aligned along a diagonal direction. This hierarchical composability enables the fine-grained vertical-vector pattern to effectively approximate diverse coarse-grained patterns, enhancing its flexibility and representational power.

\section{Extended Statistical Comparison of Different Sparse Patterns}
\label{sec:statComparison}
\begin{figure*}[t]
    \centering
    \includegraphics[width=\linewidth]{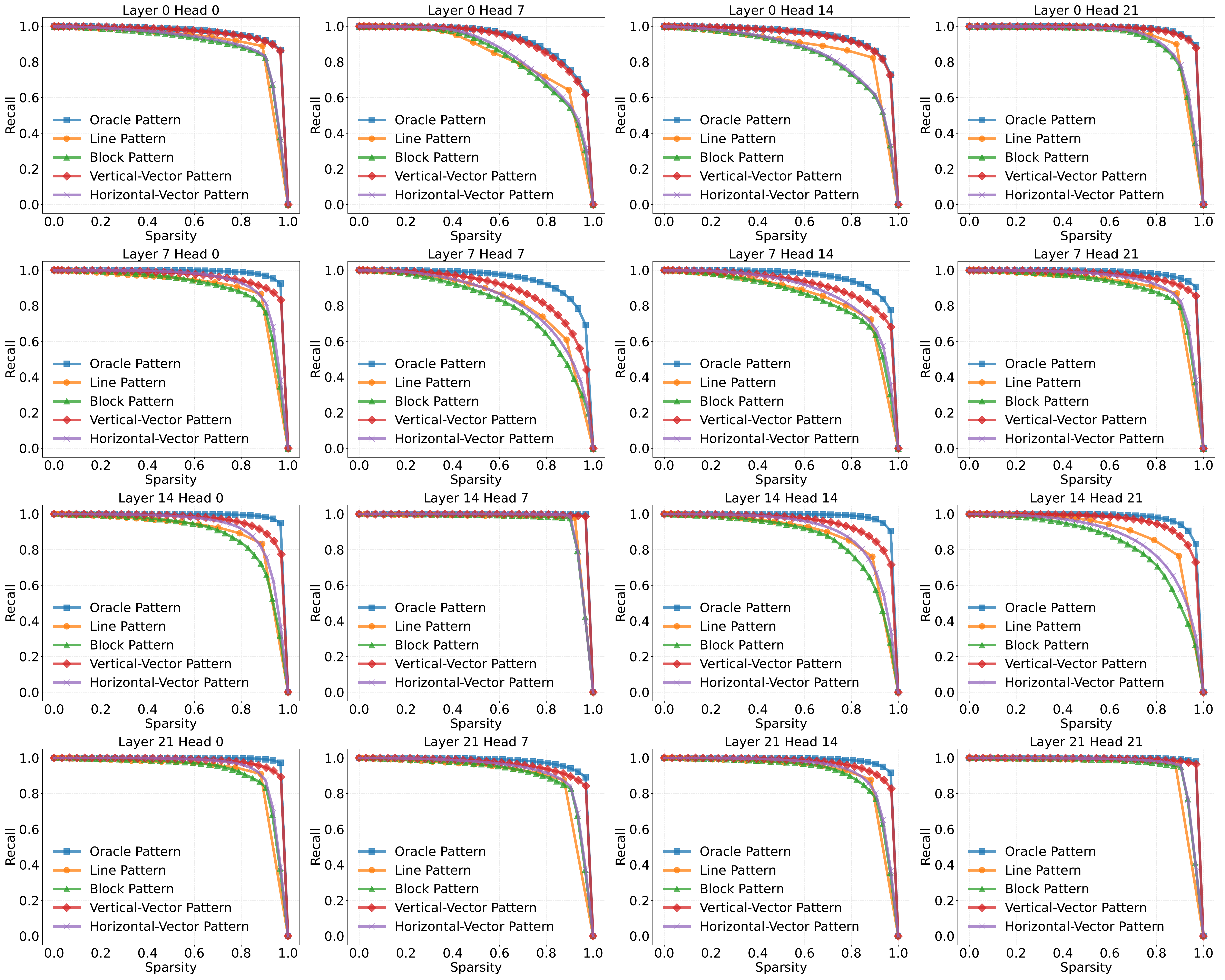}
    \caption{Sparsity-recall trade-off of various sparse patterns across different layers and attention heads of Qwen2.5-VL-7B-Instruct, evaluated on the VideoMME benchmark.}
    \label{fig:vlmmorepareto}
\end{figure*}

\begin{figure*}[t]
    \centering
    \includegraphics[width=\linewidth]{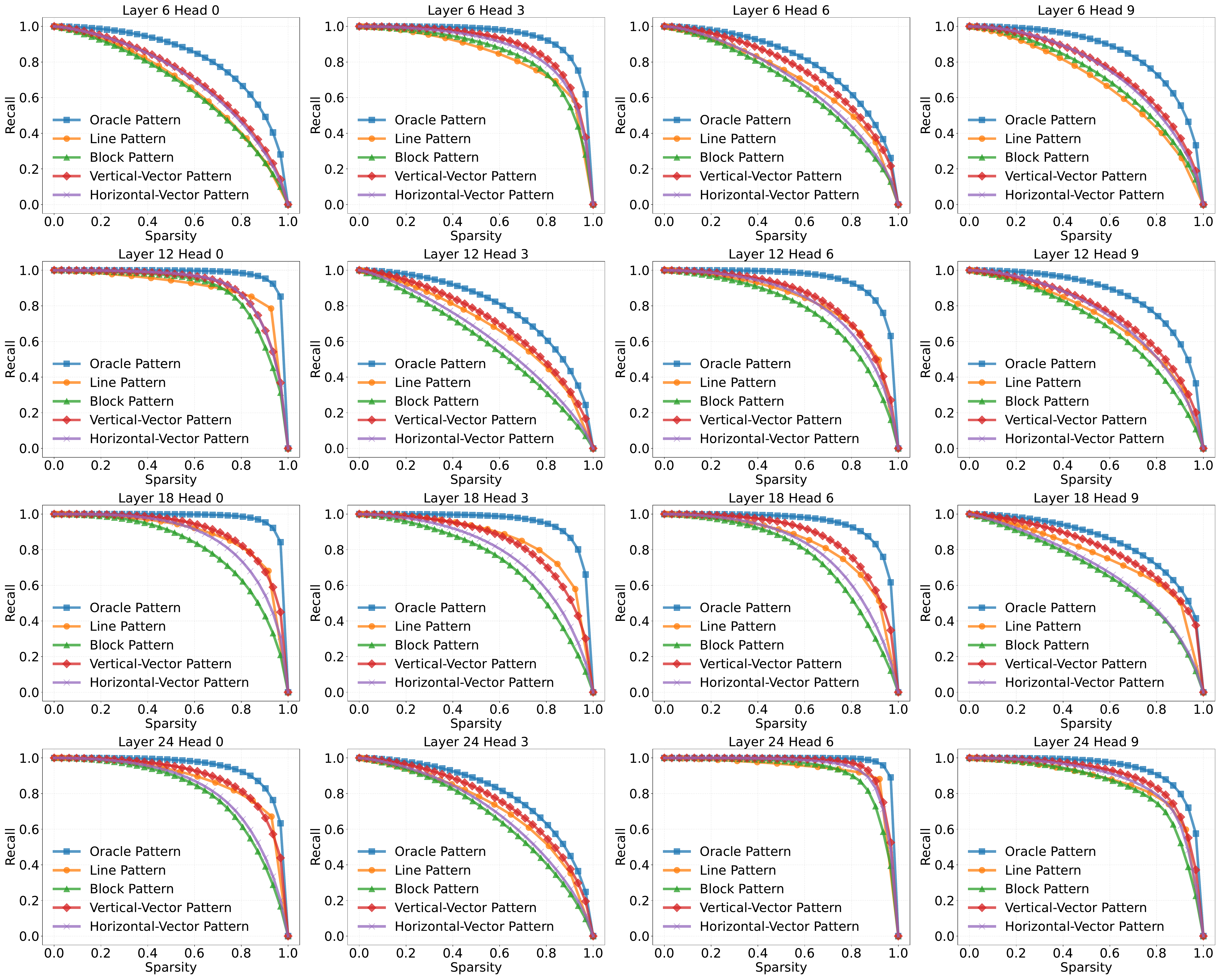}
    \caption{Sparsity-recall trade-off of various sparse patterns across different layers and attention heads of Wan2.1, evaluated on VBench.}
    \label{fig:ditmorepareto}
\end{figure*}

As shown in \Cref{fig:vlmmorepareto} and \Cref{fig:ditmorepareto}, we further approximate the attention maps presented in \Cref{fig:morevisualizationMap} using different sparse patterns, following the same approximation methodology as described in \Cref{fig:pattern_comparison_grid}. As can be observed, across nearly all attention heads, the vertical-vector pattern consistently outperforms other sparse patterns—including the fine-grained horizontal-vector pattern as well as the coarse-grained line and block patterns—achieving the closest approximation to the oracle pattern.

\section{Does Vertical-Vector Pattern Exist across Other Modalities?}
\label{sec:modality}
\begin{figure*}[ht]
    \centering
    \begin{subfigure}[b]{0.45\linewidth}
        \centering
        \includegraphics[width=\linewidth]{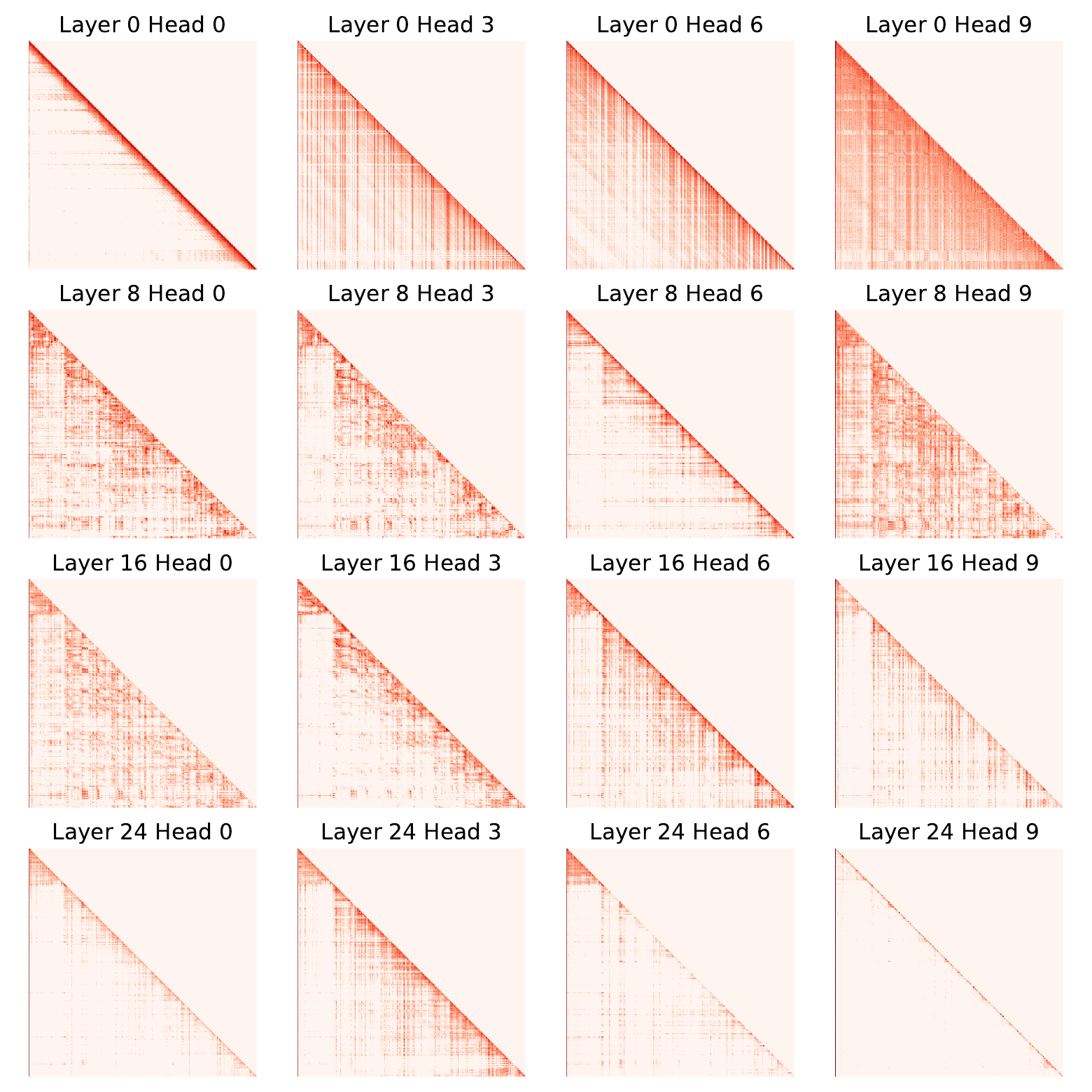}
        \caption{Attention maps}
        \label{fig:llm:a}
    \end{subfigure}
    \begin{subfigure}[b]{0.45\linewidth}
        \centering
        \includegraphics[width=\linewidth]{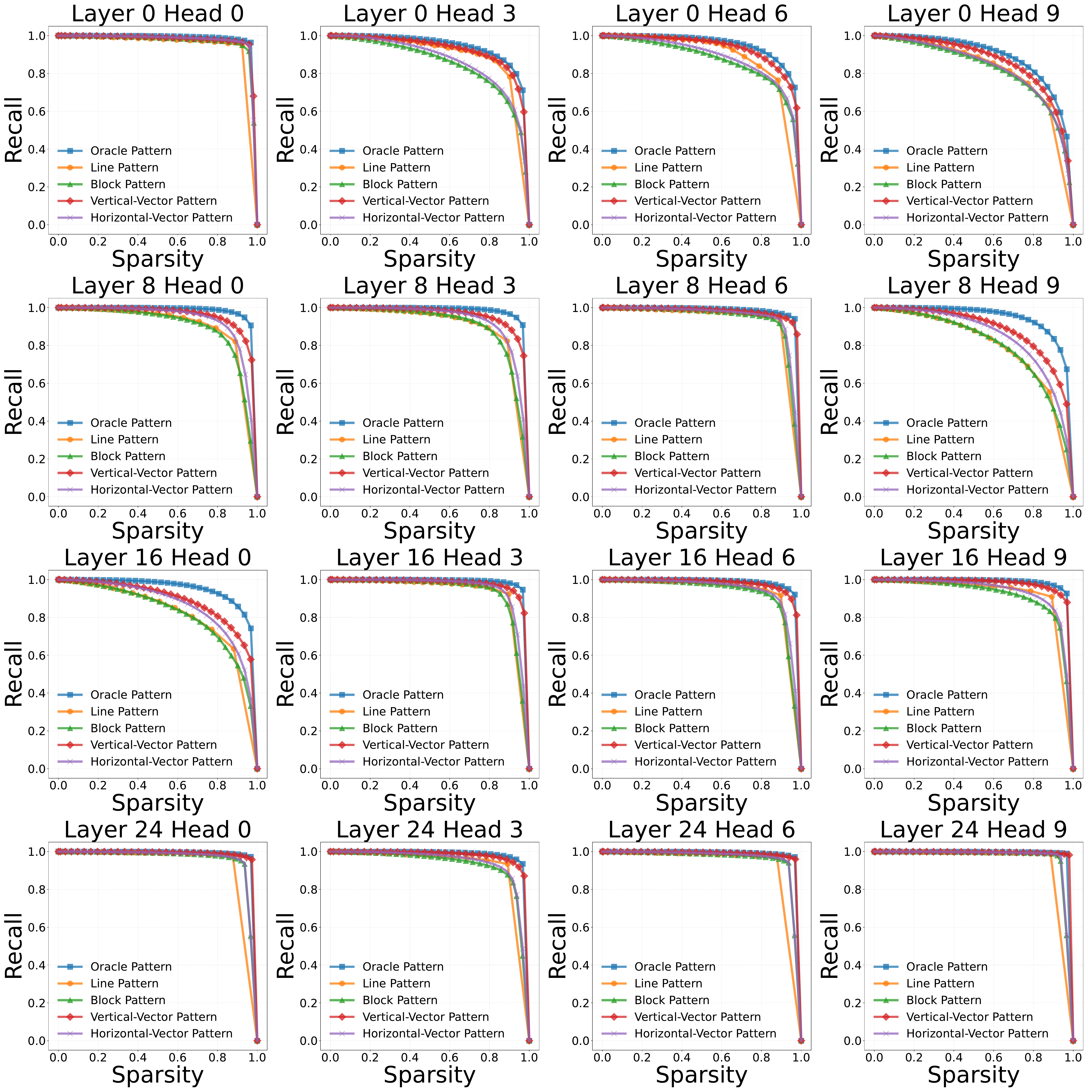}
        \caption{Sparsity-recall tradeoff of various sparse patterns}
        \label{fig:llm:b}
    \end{subfigure}
    \caption{
   The sparse pattern in Meta-Llama3-8B-Instruct evaluated on LongBench.
    }
    \label{fig:llm}
\end{figure*}

In \Cref{fig:llm:a}, we present attention maps across different layers and heads of Meta-Llama3-8B-Instruct~\cite{dubey2024llama} on LongBench~\cite{bai2024longbench}. We observe that the vertical-vector sparse pattern also exists in LLMs, albeit in a more regular and coarse-grained form—such as concentration around diagonal regions and sink tokens, or clustering within specific block-like regions. This observation is further supported by the Pareto curves of different sparse patterns in \Cref{fig:llm:b}: in many heads, the performance gap between the vertical-vector pattern and other coarse-grained patterns is small, with both closely approximating the oracle pattern.

\begin{figure*}[ht]
    \centering
    \begin{subfigure}[b]{0.45\linewidth}
        \centering
        \includegraphics[width=\linewidth]{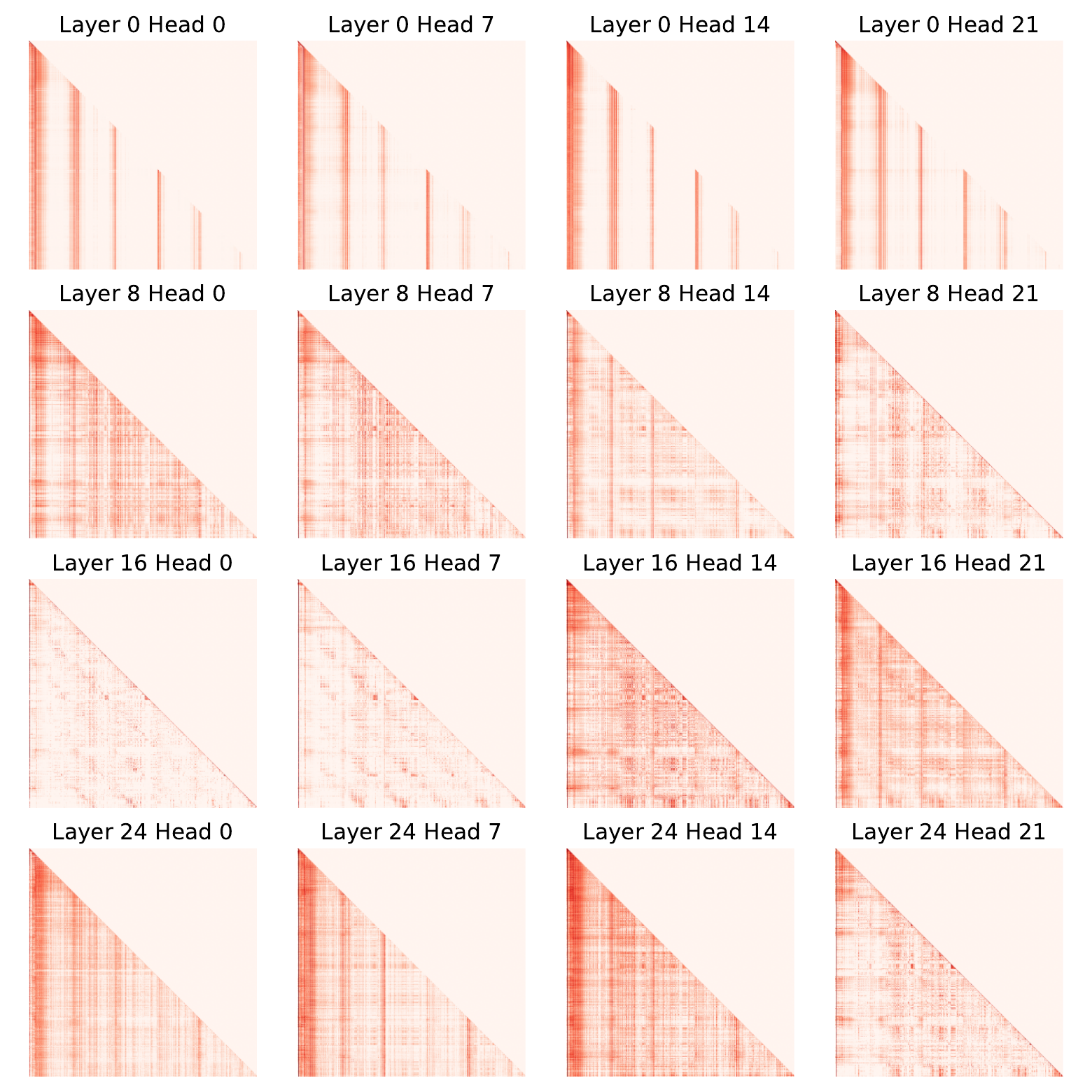}
        \caption{Attention maps}
        \label{fig:olm:a}
    \end{subfigure}
    \begin{subfigure}[b]{0.45\linewidth}
        \centering
        \includegraphics[width=\linewidth]{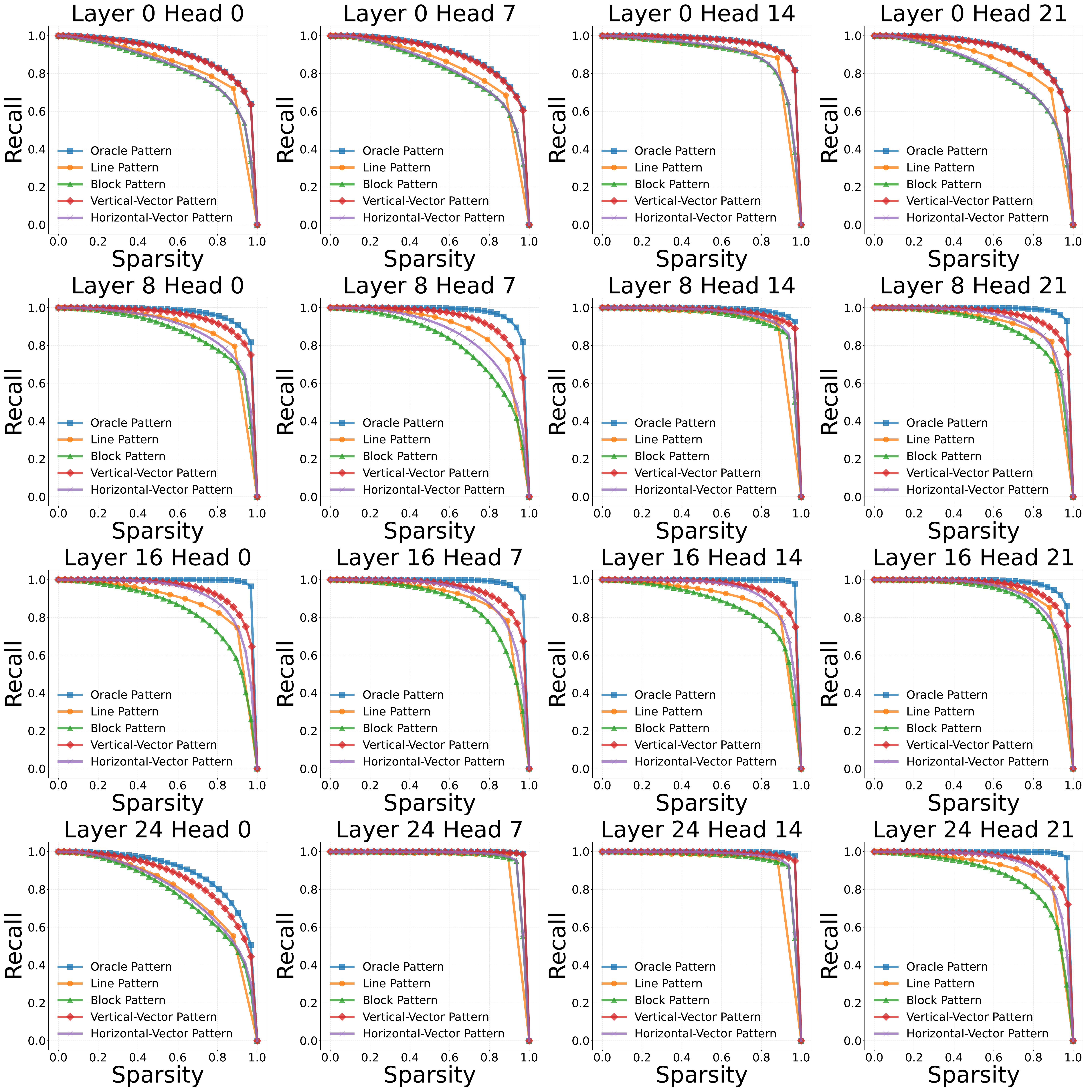}
        \caption{Sparsity-recall tradeoff of various sparse patterns}
        \label{fig:olm:b}
    \end{subfigure}
    \caption{
   The sparse pattern in Qwen2.5-Omni-7B evaluated on OmniBench~\cite{li2024omnibench}.
    }
    \label{fig:olm}
\end{figure*}

Recent studies~\cite{jiang2024minference,lai2025flexprefill,zhang2025spargeattn} have identified such regular, coarse-grained sparse patterns in LLMs, and leveraged them for model acceleration—e.g., FlexPrefill~\cite{lai2025flexprefill} exploiting vertical-slash patterns and XAttention~\cite{xu2025xattention} utilizing block patterns. More recently, AnchorAttention~\cite{zhang2025anchorattention} explicitly identifies anchor tokens in LLMs to exploit vertical-vector sparsity, demonstrating that the vertical-vector pattern is not only prevalent but also highly promising for practical efficiency improvements. Furthermore, the vertical-vector sparse pattern is also observed in Qwen2.5-Omni-7B~\cite{xu2025qwen25omnitechnicalreport}, which jointly processes text, images, and audio, as shown in \Cref{fig:olm}, demonstrating its practical potential even under more complex, multimodal input scenarios.

\section{Kernel Implementations}
\label{sec:kernel}
\subsection{Important-Vector Selection}
As shown in \Cref{alg:important_vector_selection}, our kernel of Important-Vector Selection provides an efficient approximation of the most informative key vectors by combining the \textit{TilingSelect} strategy with a lightweight \textit{minS} filtering mechanism. 
The kernel operates at the granularity of query--key tiles: each GPU thread block is responsible for one query tile of size $B_Q$ and a group of $G_K$ key tiles, and performs tiling-based GEMM computations with the on-the-fly filtering to identify high-saliency key positions.

During the iteration over K-tiles, the kernel maintains a \textit{per-row running maximum}---a row-wise accumulator stored in on-chip memory---which records, for each query row, the maximum attention score observed across all previously processed tiles. 
After each tile’s partial attention matrix is computed, its row-wise maximum is compared against the accumulator to update a progressively tighter global threshold. 
This accumulated maximum is then reused by subsequent tiles to perform \textit{cross-tile minS filtering}, enabling more stable and global importance estimation without revisiting previous tiles or re-evaluating thresholds.

For each query tile, the kernel extracts both  
(i) an \textit{important-vector index set}, formed by concatenating all selected indices returned from tile-level filtering, and  
(ii) a \textit{per-row count vector}, obtained by summing the number of selected key positions contributed by each tile.  
These two outputs correspond to the index matrix $\boldsymbol{I} \in \mathbb{N}^{N_p \times n_k}$, and the count vector $\boldsymbol{C} \in \mathbb{N}^{N_p}$ in~\Cref{alg:important_vector_selection}. Here $N_p$ is the length of $\boldsymbol{Q}_p$ and $n_k=\max_{1\le j \le N_p} \boldsymbol{C}_j$, 

By coupling tile-structured GEMM, cross-tile max accumulation, and cross-tile minS filtering, the kernel yields a compact and accurate approximation of the important key vectors with minimal memory traffic and computational overhead. 
The resulting index and count sets $(\boldsymbol{I}, \boldsymbol{C})$ are subsequently consumed by the downstream kernel of vector-sparse attention.

\subsection{Vector-Sparse Computation}

\Cref{alg:v_attn} presents the implementation of our Vector-Sparse Attention Computation kernel. 
The kernel follows a similar design spirit to the kernel of vertical-slash sparse FlashAttention in MInference~\cite{jiang2024minference}, 
but operates without block counts or block-level indexing. 
Instead, it directly consumes the \emph{important-vector index matrix} $\boldsymbol{I}$ 
and the corresponding \emph{important-vector count vector} $\boldsymbol{C}$ 
generated by the upstream kernel of Important-Vector Selection in \Cref{alg:important_vector_selection}.

In the kernel, each thread block is assigned to process a single Q-tile, whose size is set to 
$B_Q = P_q$, matching the vector size used during important-vector selection. 
This choice is intentional: 
under the vertical-vector sparse pattern, all queries within a vector-sized region 
share the same set of important key--value rows. 
Thus, assigning one Q-tile of size $B_Q = P_q$ per thread block yields shared 
index access patterns.
For each Q-tile, the kernel iterates over the important vertical vectors specified by 
$\boldsymbol{I}$ and $\boldsymbol{C}$, and loads only the required K/V rows via 
$\operatorname{GatherLoad}$. 
It then performs tile-level attention score computation followed by FlashAttention-style 
running-max and running-normalizer accumulation. 
These accumulators, $\boldsymbol{m}^{(i_q)}$ and $\boldsymbol{\ell}^{(i_q)}$, 
are maintained in on-chip memory and updated incrementally as each important-Key tile is processed. 

The kernel’s latency is directly determined by the sparsity structure discovered 
during Important-Vector Selection, allowing the model to benefit from large reductions 
in attention computation when the attention map contains high sparsity.

\section{Comparison across Different Context Lengths}
\begin{figure}[H]
    \centering
    \includegraphics[width=\linewidth]{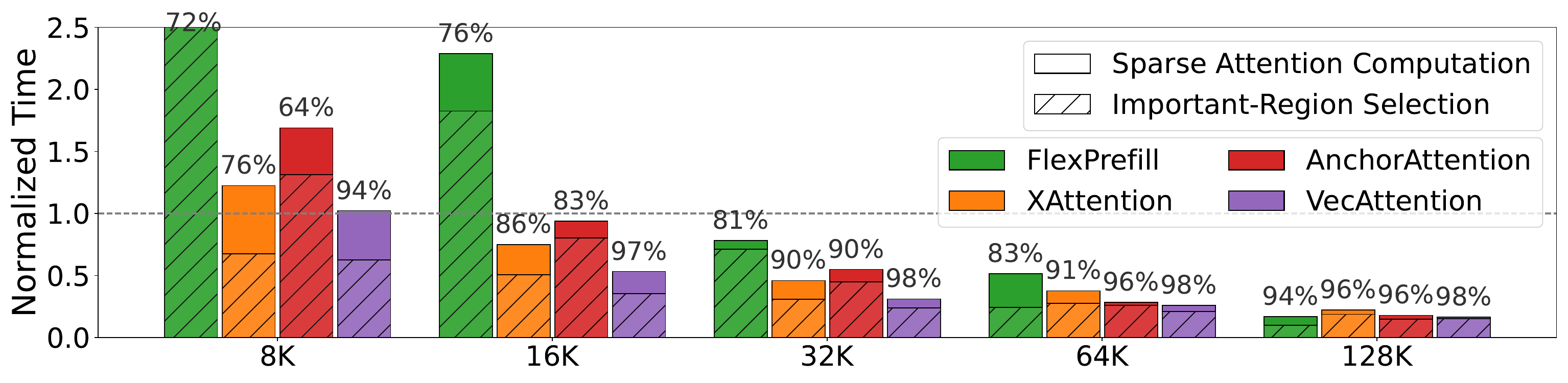}
    \caption{Attention latency breakdown normalized to full attention across different context lengths on V-NIAH~\cite{zhang2024long}. Markers (e.g., 64\%) indicate the sparsity ratio.} %
    \label{fig:attn_len}
\end{figure}
As shown in~\Cref{fig:attn_len}, \name{} consistently outperforms all baselines in terms of attention latency across context lengths ranging from 8K to 128K. As the input context length increases, \name{} leverages the fine-grained vertical-vector sparse pattern to achieve higher sparsity ratios without compromising accuracy, thereby significantly reducing sparse attention computation time. Moreover, the TilingSelect and minS techniques in \name{} effectively address the performance bottleneck in fine-grained vector selection (as detailed in~\Cref{sec:imp_vec_select}), resulting in a selection overhead that is comparable to—or lower than—that of coarse-grained baselines.

\section{Qualitative Comparison of Videos Generated by Different Methods}
\label{sec:qualitative}
\begin{figure*}[t]
    \centering
    \begin{subfigure}[b]{\textwidth}
        \centering
        \includegraphics[width=\textwidth]{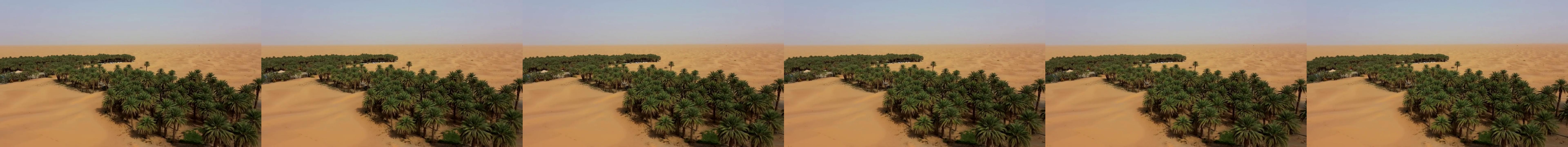}
        \caption{Full Attention}
    \end{subfigure}
    \begin{subfigure}[b]{\textwidth}
        \centering
        \includegraphics[width=\textwidth]{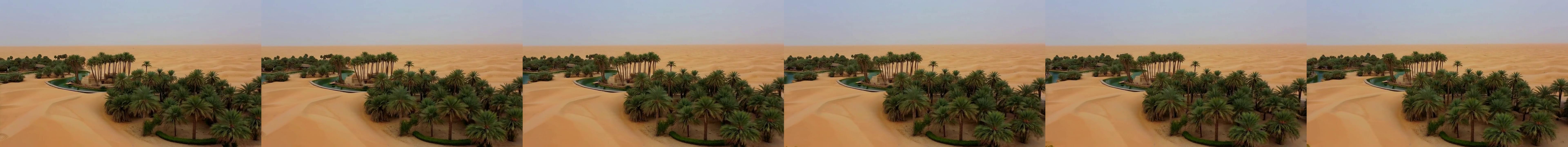}
        \caption{XAttention}
    \end{subfigure}
    \begin{subfigure}[b]{\textwidth}
        \centering
        \includegraphics[width=\textwidth]{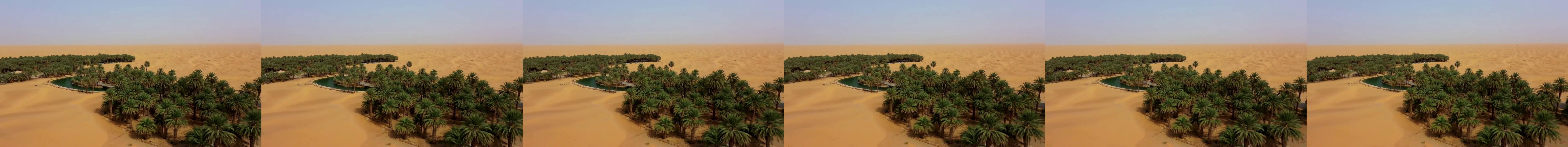}
        \caption{Sparse VideoGen}
    \end{subfigure}
    \begin{subfigure}[b]{\textwidth}
        \centering
        \includegraphics[width=\textwidth]{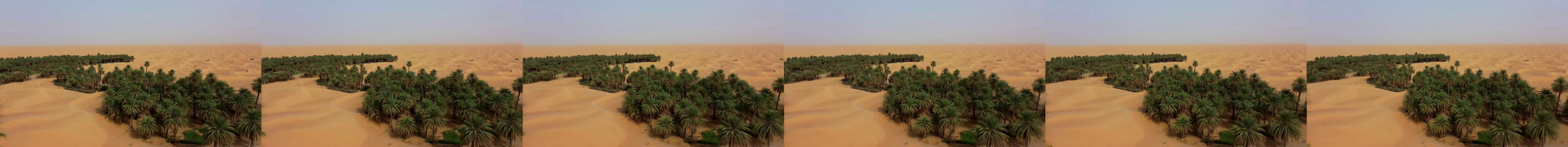}
        \caption{VecAttention}
    \end{subfigure}
    \caption{Qualitative comparison of videos generated with different sparse attention methods. The input prompt is sampled from VBench: ``In the vast desert, an aerial shot shows an oasis nestled among the dunes, featuring tall palm trees and exuding an air of serenity. Realistic, Natural lighting, Peaceful.''}
    \label{fig:videogen}
\end{figure*}

As shown in \Cref{fig:videogen}, the video generated by \name{} is the closest to that produced with Full Attention. It successfully preserves both the visual elements and temporal consistency characteristic of Full Attention generation. In contrast, the baseline methods—XAttention and Sparse VideoGen—introduce extraneous artifacts, such as additional bodies of water within the oasis, not described in the prompt. This comparison demonstrates that \name{} more accurately identifies fine-grained, semantically important regions for attention, effectively maintaining the fidelity and precision of the video generation model without compromising quality.

\section{Limitations}
\label{sec:limitation}
Our work has two main limitations, which also point to promising directions for future research. First, while we have demonstrated the effectiveness of the vertical-vector sparse pattern across different modalities, its generalization to complex downstream tasks remains to be fully explored. For instance, evaluation on applications in dynamic scenarios such as agent reasoning~\cite{wu2025agentic} or retrieval-augmented generation (RAG)~\cite{arslan2024survey} is still limited. As a next step, we plan to extend our analysis across a more diverse set of models and downstream tasks, with the goal of identifying modality- and task-adaptive sparsity strategies.
Second, our current study primarily focuses on vertical-vector patterns, leaving other fine-grained structures—such as horizontal vectors or diagonal (slash) vectors—less examined in terms of their end-to-end impact on accuracy and efficiency. While initial results suggest that vertical structures are dominant in many cases, different patterns may offer complementary advantages under specific conditions. In future work, we aim to develop a more comprehensive framework that dynamically selects or combines multiple fine-grained sparse patterns based on input characteristics and model behavior.

\restylefloat{algorithm}

\begin{algorithm}[htbp]
\caption{Important-Vector Selection}
\label{alg:important_vector_selection}
\begin{algorithmic}[1]

  \STATE \textbf{Input:} 
    $\boldsymbol{Q}, \boldsymbol{K} \in \mathbb{R}^{N \times D}$; 
    query pooling size $P_q$; 
    K-tile size $B_K$; 
    K-tile-group size $G_K$; 
    minS filtering ratio $\alpha$
  \STATE \textbf{Output:} 
    important-vector index $\boldsymbol{I} \in \mathbb{N}^{N_p \times n_k}$; count $\boldsymbol{C} \in \mathbb{N}^{N_p}$

  \LineComment{Row-wise mean pooling on queries (\Cref{eq:query_pooling})}
  \STATE $\boldsymbol{Q}_p \gets \operatorname{MeanPool}(\boldsymbol{Q}, P_q) 
          \in \mathbb{R}^{N_p \times D}$

  \LineComment{Number of Q-tiles and K-tile-groups}
  \STATE $N_q \gets \left\lceil \frac{N_p}{B_Q} \right\rceil$, \quad
         $N_g \gets \left\lceil \frac{N}{B_K G_K} \right\rceil$

  \ParallelComment{Parallel across Q-tile dimension}
  \FOR{$i_q = 0$ \textbf{to} $N_q - 1$}
      \STATE $\boldsymbol{Q}^{(i_q)} 
      \gets \boldsymbol{Q}_p[i_q B_Q : (i_q+1) B_Q, :]
      \in \mathbb{R}^{B_Q \times D}$

      \STATE $\boldsymbol{I}^{(i_q)} \gets [\,]$ \LineComment{local index buffer}
      \STATE $\boldsymbol{C}^{(i_q)} \gets \boldsymbol{0}_{B_Q}
              \in \mathbb{R}^{B_Q}$ \LineComment{per-row count buffer}

      \ParallelComment{Parallel across K-tile-groups}
      \FOR{$i_g = 0$ \textbf{to} $N_g - 1$}
          \STATE $\delta_k \gets i_g \cdot B_K G_K$

          \LineComment{Collect K-tiles in this group}
          \STATE $\mathcal{K}^{(i_g)} 
          \gets \{\boldsymbol{K}[\delta_k + j B_K : \delta_k + (j+1) B_K, :]
               \mid j = 0,\ldots,G_K-1\}$

          \STATE $\boldsymbol{m}_S \gets (-\infty)\,\boldsymbol{1}_{B_Q}
          \in \mathbb{R}^{B_Q}$

          \LineComment{Iterate over K-tiles within the group}
          \FOR{$\boldsymbol{K}^{\text{tile}} \in \mathcal{K}^{(i_g)}$}

              \STATE $\boldsymbol{S}^{\text{tile}} 
              \gets \boldsymbol{Q}^{(i_q)} (\boldsymbol{K}^{\text{tile}})^\top / \sqrt{D}  
              \in \mathbb{R}^{B_Q \times B_K}$

              \LineComment{Update row-wise maximum scores}
              \STATE $\boldsymbol{m}_S 
              \gets \operatorname{max}\!\left(
                   \boldsymbol{m}_S,\,
                   \operatorname{rowmax}(\boldsymbol{S}^{\text{tile}})
                 \right)
                 \in \mathbb{R}^{B_Q}$

              \LineComment{Binary mask using minS filtering (\Cref{eqn:idx})}
              \STATE $\boldsymbol{M}
              \gets \boldsymbol{S}^{\text{tile}} > (\boldsymbol{m}_S-\alpha)$

              \LineComment{Selected indices and per-row counts for this tile}
              \STATE $(\boldsymbol{I}^{\text{tile}},\, \boldsymbol{C}^{\text{tile}})
                  \gets \operatorname{Indexing}(\boldsymbol{M},\, \delta_k)$

              \LineComment{Accumulate indices and counts}
              \STATE $\boldsymbol{I}^{(i_q)} 
              \gets \operatorname{Concatenate}\!\big(
                   \boldsymbol{I}^{(i_q)},\, \boldsymbol{I}^{\text{tile}}
                  \big)$
              \STATE $\boldsymbol{C}^{(i_q)} 
              \gets \boldsymbol{C}^{(i_q)} + \boldsymbol{C}^{\text{tile}}$

              \LineComment{Move K offset to next K-tile}
              \STATE $\delta_k \gets \delta_k + B_K$

          \ENDFOR

      \ENDFOR

  \ENDFOR

  \LineComment{Merge all index and count of each Q-tile}
  \STATE $\boldsymbol{I} 
    \gets \operatorname{Concatenate}\!\big(
         \boldsymbol{I}^{(0)}, \boldsymbol{I}^{(1)}, \ldots, \boldsymbol{I}^{(N_q-1)}
       \big)$
  \STATE $\boldsymbol{C} 
    \gets \operatorname{Concatenate}\!\big(
         \boldsymbol{C}^{(0)}, \boldsymbol{C}^{(1)}, \ldots, \boldsymbol{C}^{(N_q-1)}
       \big)$

\end{algorithmic}
\end{algorithm}

\restylefloat{algorithm}

\restylefloat{algorithm}

\begin{algorithm}[htbp]
\caption{Vector-Sparse Computation}
\label{alg:v_attn}
\begin{algorithmic}[1]

  \STATE \textbf{Input:}
    $\boldsymbol{Q},\boldsymbol{K},\boldsymbol{V} \in \mathbb{R}^{N \times D}$;
    vector size $P_q$ (we set the Q-tile size $B_Q = P_q$);
    important-vector index $\boldsymbol{I} \in \mathbb{N}^{N_p \times n_k}$;
    important-vector count $\boldsymbol{C} \in \mathbb{N}^{N_p}$;
  \STATE \textbf{Output:}
    attention output $\boldsymbol{O} \in \mathbb{R}^{N \times D}$

  \STATE Scale factor $\tau \gets 1 / \sqrt{D}$
  \STATE Initialize $\boldsymbol{O} \gets \boldsymbol{0}_{N \times D}
         \in \mathbb{R}^{N \times D}$

  \LineComment{Number of Q-tiles (tile size $B_Q = P_q$)}
  \STATE $N_q \gets \left\lceil \frac{N}{B_Q} \right\rceil$

  \ParallelComment{Parallel across Q-tile dimension}
  \FOR{$i_q = 0$ \textbf{to} $N_q - 1$}

      \STATE $\boldsymbol{Q}^{(i_q)} 
      \gets \boldsymbol{Q}[i_q B_Q : (i_q+1) B_Q, :]
      \in \mathbb{R}^{B_Q \times D}$

      \LineComment{Initialize tile-local accumulators}
      \STATE $\boldsymbol{O}^{(i_q)} \gets \boldsymbol{0}_{B_Q \times D}
              \in \mathbb{R}^{B_Q \times D}$
      \STATE $\boldsymbol{m}^{(i_q)} \gets (-\infty)\,\boldsymbol{1}_{B_Q}
              \in \mathbb{R}^{B_Q}$
      \STATE $\boldsymbol{\ell}^{(i_q)} \gets \boldsymbol{0}_{B_Q}
              \in \mathbb{R}^{B_Q}$

      \LineComment{Loop over vertical-vector indices}
      \STATE $j \gets 0$
      \WHILE{$j < \boldsymbol{C}[i_q]$}
          \STATE $\boldsymbol{i}^{(i_q,j)} 
          \gets \boldsymbol{I}[i_q,\; j : j + B_K]
          \in \mathbb{N}^{B_K}$

          \LineComment{Load K/V rows referenced by $\boldsymbol{i}^{(i_q,j)}$}
          \STATE $\boldsymbol{K}^{\text{tile}} 
          \gets \operatorname{GatherLoad}(\boldsymbol{K}, \boldsymbol{i}^{(i_q,j)})
          \in \mathbb{R}^{B_K \times D}$
          \STATE $\boldsymbol{V}^{\text{tile}} 
          \gets \operatorname{GatherLoad}(\boldsymbol{V}, \boldsymbol{i}^{(i_q,j)})
          \in \mathbb{R}^{B_K \times D}$

          \STATE $\boldsymbol{S} 
          \gets \tau\, \boldsymbol{Q}^{(i_q)} (\boldsymbol{K}^{\text{tile}})^\top
          \in \mathbb{R}^{B_Q \times B_K}$
          \STATE $\boldsymbol{S} \gets \operatorname{mask}(\boldsymbol{S})$

          \STATE $\boldsymbol{m}^{(i_q)}_{\text{new}} 
          \gets \operatorname{max}\!\big(
               \boldsymbol{m}^{(i_q)},\;
               \operatorname{rowmax}(\boldsymbol{S})
             \big)
             \in \mathbb{R}^{B_Q}$

          \STATE $\boldsymbol{S} \gets 
            \boldsymbol{S} - \boldsymbol{m}^{(i_q)}_{\text{new}}$

          \STATE $\boldsymbol{P} \gets \exp(\boldsymbol{S})
                  \in \mathbb{R}^{B_Q \times B_K}$
          \STATE $\boldsymbol{\ell}^{(i_q)}_{\text{new}} 
          \gets \operatorname{rowsum}(\boldsymbol{P})
             \in \mathbb{R}^{B_Q}$

          \STATE $\boldsymbol{\alpha} 
          \gets \exp\!\big(\boldsymbol{m}^{(i_q)} - \boldsymbol{m}^{(i_q)}_{\text{new}}\big)
             \in \mathbb{R}^{B_Q}$

          \STATE $\boldsymbol{\ell}^{(i_q)} 
          \gets \boldsymbol{\alpha} \boldsymbol{\ell}^{(i_q)}
               + \boldsymbol{\ell}^{(i_q)}_{\text{new}}$
          \STATE $\boldsymbol{O}^{(i_q)} 
          \gets \boldsymbol{\alpha} \boldsymbol{O}^{(i_q)}
               + \boldsymbol{P}\boldsymbol{V}^{\text{tile}}$

          \STATE $\boldsymbol{m}^{(i_q)} \gets \boldsymbol{m}^{(i_q)}_{\text{new}}$

          \STATE $j \gets j + B_K$
      \ENDWHILE

      \STATE $\boldsymbol{O}^{(i_q)} 
      \gets \operatorname{diag}\!\big(\boldsymbol{\ell}^{(i_q)}\big)^{-1}
           \boldsymbol{O}^{(i_q)}$

      \STATE $\boldsymbol{O}[i_q B_Q : (i_q+1) B_Q, :]
      \gets \boldsymbol{O}^{(i_q)}$

  \ENDFOR

\end{algorithmic}
\end{algorithm}

\end{document}